\documentclass[11pt]{article}

\usepackage[final]{acl}

\usepackage{times}
\usepackage{latexsym}

\usepackage[T1]{fontenc}

\usepackage[utf8]{inputenc}

\usepackage{microtype}

\usepackage{inconsolata}

\usepackage{graphicx}

\usepackage{color,xcolor}
\usepackage{epsfig}
\usepackage{graphicx}
\usepackage{tabularx}
\usepackage{times}
\usepackage{latexsym}

\usepackage{adjustbox}
\usepackage{array}
\usepackage{booktabs}
\usepackage{colortbl}
\usepackage{float,wrapfig}
\usepackage{hhline}
\usepackage{multirow}
\usepackage{subcaption} 
\usepackage[toc,page]{appendix}

\usepackage{amsmath,amsfonts,amsthm,amssymb}
\usepackage{bm}
\usepackage{nicefrac}
\usepackage{microtype}
\usepackage{inconsolata}
\usepackage{pifont}

\usepackage{changepage}
\usepackage{cuted}
\usepackage{extramarks}
\usepackage{fancyhdr}
\usepackage{lastpage}
\usepackage{setspace}
\usepackage{soul}
\usepackage{xspace}
\usepackage{paralist}
\usepackage{booktabs,arydshln}

\usepackage{hyperref}
\hypersetup{
  urlcolor=mydarkblue
}
\usepackage{url}

\usepackage{algorithm, algpseudocode}
\usepackage{enumerate}
\usepackage{lipsum}
\usepackage{physics}
\usepackage{cleveref}

\usepackage{listings}
\usepackage{xcolor}
\usepackage{enumitem}

\definecolor{commentgray}{rgb}{0.5, 0.5, 0.5}

\newcolumntype{L}[1]{>{\raggedright\let\newline\\\arraybackslash\hspace{0pt}}m{#1}}
\newcolumntype{C}[1]{>{\centering\let\newline\\\arraybackslash\hspace{0pt}}m{#1}}
\newcolumntype{R}[1]{>{\raggedleft\let\newline\\\arraybackslash\hspace{0pt}}m{#1}}

\newcommand{\ignorethis}[1]{}

\makeatletter
\DeclareRobustCommand\onedot{\futurelet\@let@token\@onedot}
\def\@onedot{\ifx\@let@token.\else.\null\fi\xspace}

\makeatother

\makeatletter
\def\adl@drawiv#1#2#3{%
        \hskip.5\tabcolsep
        \xleaders#3{#2.5\@tempdimb #1{1}#2.5\@tempdimb}%
                #2\z@ plus1fil minus1fil\relax
        \hskip.5\tabcolsep}
\newcommand{\cdashlinelr}[1]{%
  \noalign{\vskip\aboverulesep
           \global\let\@dashdrawstore\adl@draw
           \global\let\adl@draw\adl@drawiv}
  \cdashline{#1}
  \noalign{\global\let\adl@draw\@dashdrawstore
           \vskip\belowrulesep}}
\makeatother

\definecolor{tablepink}{RGB}{250, 235, 242}
\definecolor{tableblue}{RGB}{210, 225, 245}
\definecolor{mydarkblue}{rgb}{0,0.08,1}
\definecolor{mydarkgreen}{rgb}{0.02,0.6,0.02}
\definecolor{mydarkred}{rgb}{0.8,0.02,0.02}
\definecolor{mydarkorange}{rgb}{0.40,0.2,0.02}
\definecolor{mypurple}{RGB}{111,0,255}
\definecolor{myred}{rgb}{1.0,0.0,0.0}
\definecolor{mygold}{rgb}{0.75,0.6,0.12}
\definecolor{mydarkgray}{rgb}{0.66, 0.66, 0.66}

\definecolor{darkgreen}{rgb}{0.02,0.6,0.02}
\definecolor{darkred}{rgb}{0.8,0.02,0.02}
\definecolor{darkorange}{rgb}{0.40,0.2,0.02}
\definecolor{darkpurple}{RGB}{111,0,255}

\theoremstyle{plain}

\theoremstyle{definition}

\theoremstyle{remark}

\title{GroupDPO: Memory-Efficient Group-Wise Direct Preference Optimization}

\author{
Jixuan Leng\thanks{Work done while interning at Google.} \\ 
Carnegie Mellon University \\
\texttt{jixuanl@cs.cmu.edu} \\
\And
Si Si\\
Google DeepMind \\
\texttt{sisidaisy@google.com} \\
\AND
Hsiang-Fu Yu \\
Google\\
\texttt{rofuyu@google.com} \\
\And
Vinod Raman \\
Google DeepMind\\
\texttt{vinodraman@google.com} \\
\And
Inderjit S. Dhillon\\
Google\\
\texttt{isd@google.com} \\
}

\begin{document}
\maketitle
\begin{abstract}
Preference optimization is widely used to align Large Language Models (LLMs) with preference feedback.
However, most existing methods train on a single positive-negative pair per prompt, discarding additional supervision available in preference datasets that typically contain multiple candidate responses.
Motivated by this limitation, recent work explores group-wise preference optimization, which jointly contrasts multiple responses for the same prompt, but its empirical behavior and scalability remain underexplored due to the memory overhead of group-coupled objectives.
In this work, we present a unified empirical and systems study of group-wise preference optimization and develop a memory-efficient implementation for group-coupled objectives.
By instantiating first-order linearization with objective-specific per-response coefficients, our implementation preserves first-order gradients while decoupling samples during backpropagation, substantially reducing peak memory usage and enabling scalable training with larger groups.
Across offline and online settings, we show that leveraging multiple responses consistently outperforms single-pair training. 
Furthermore, incorporating a negative log-likelihood (NLL) term on positive responses is critical for both performance gains and training stability.
\end{abstract}

\section{Introduction}
\vspace{-0.2em}
\label{sec:intro}
\begin{figure*}[ht]
  \centering
\includegraphics[width=\textwidth]{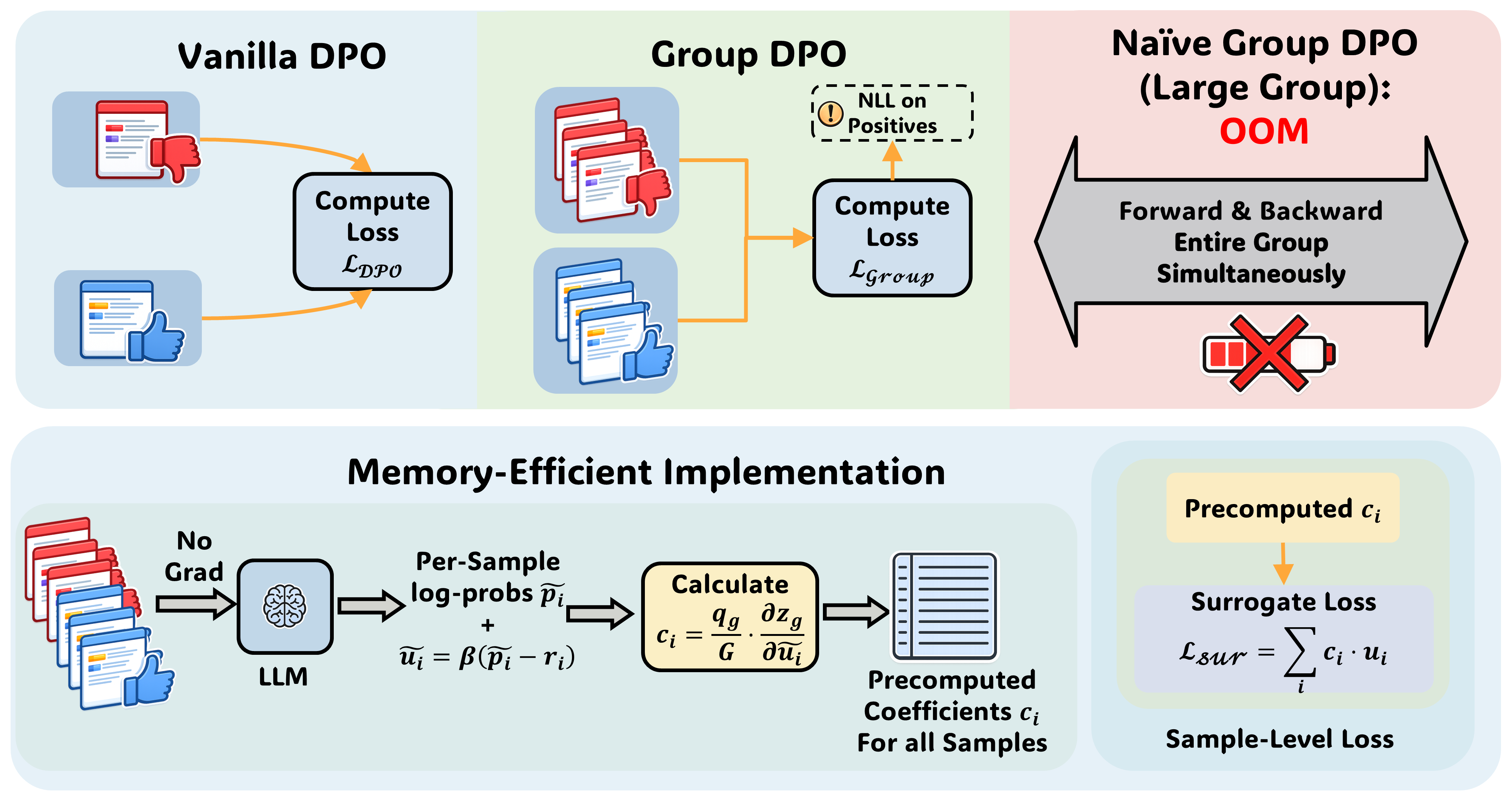}
\caption{Overview of Group DPO and our memory-efficient surrogate implementation. 
Top: Standard DPO trains on a single positive-negative pair, while group-wise objectives leverage multiple responses but require joint forward and backward passes, leading to high memory usage and OOM with large group sizes. 
Bottom: Our surrogate avoids group-coupled backpropagation by precomputing per-sample coefficients in a no-gradient pass and optimizing a sample-level loss, substantially reducing peak memory overhead and enabling scalable training with larger groups.}
\label{fig:overview}
\vspace{-0.5em}
\end{figure*}

Preference optimization has become a central component of modern large language model (LLM) alignment.
Rather than relying solely on supervised fine-tuning, many alignment pipelines train policy models using preference feedback that compares alternative responses to the same prompt.
Early approaches relied on Reinforcement Learning from Human Feedback (RLHF)~\citep{christiano2017deep, ouyang2022training, yang2025reliable}, where a reward model is first trained then used to optimize the policy model.
More recently, Direct Preference Optimization (DPO)~\citep{rafailov2023direct} and related methods~\citep{meng2024simpo, hong2024orpo, liu2024provably} simplify this pipeline by directly optimizing the policy from preference data, bypassing reward-model training and policy-gradient RL.

Despite these advances, most preference optimization methods operate on \emph{pairwise} comparisons between a single positive and negative response.
In practice, however, preference datasets are often constructed by sampling multiple candidate responses for each prompt and evaluating them using human or automated feedback~\citep{cui2023ultrafeedback}.
To apply pairwise objectives such as DPO, these response sets are typically \emph{binarized} by selecting a single positive and negative response while discarding the remaining candidates.
This reduction loses potentially useful supervision signals about the relative quality of responses within the group.
As a result, recent work has begun exploring \emph{group-wise} or \emph{listwise} preference learning methods that train on response sets rather than individual pairs.

Group-wise preference optimization can provide richer supervision by jointly contrasting multiple candidate responses for the same prompt.
Several recent methods extend pairwise preference learning to this setting using ranking or set-level objectives~\citep{song2024preference, gupta2024multi, liu2025lipo, chen2024softmax}.
However, the practical behavior of these methods remains poorly understood, as prior work typically proposes individual objectives and evaluates them in isolation, making systematic comparisons under consistent training settings difficult.
Moreover, group-wise objectives introduce cross-response dependencies within each prompt group.
Because the loss depends on interactions between positive and negative responses, gradients become coupled across samples and require constructing a joint computation graph over the group.
In naive implementations~\citep{vonwerra2020trl,gupta2024multi}, activations for all samples in a group must therefore be retained until backpropagation, causing activation memory to scale with group size and quickly become prohibitive.
Consequently, memory costs restrict prior work to relatively small group sizes, limiting exploitation of group-level supervision.

In this work, we study group-wise preference optimization from both practical and empirical perspectives.
To address the computational challenges of group-wise objectives, we develop a memory-efficient surrogate implementation that reformulates the objective into a sample-level loss with matched first-order gradients.
Our construction instantiates a general first-order linearization principle for group-wise preference objectives: we derive objective-specific per-sample coefficients using a lightweight no-gradient forward pass and then perform standard token-level backpropagation with these coefficients.
This decouples gradients across samples and eliminates the need to retain activations for samples simultaneously, making memory overhead largely independent of the group size.

Using this implementation, we unify and compare several representative group-wise objectives from prior work under controlled offline and online alignment settings.
Our results reveal two robust design principles.
First, incorporating multiple samples per prompt consistently outperforms single-pair training, indicating that group-level supervision provides a richer learning signal.
Second, group-wise training can be unstable: including an additional negative log-likelihood (NLL) term~\citep{liu2024provably, pang2024iterative, wang2024preference, grattafiori2024llama, pal2024smaug} over positive samples plays a crucial role in preventing training collapse and improving the final performance.

We evaluate our approach across multiple models and training settings.
Results on multi-domain benchmarks show that group-wise preference optimization consistently outperforms single-pair training.
Meanwhile, the proposed surrogate implementation significantly reduces memory usage while maintaining competitive training latency, making group-wise training more practical at larger scales.

Our contributions are summarized as follows:
\begin{itemize}[leftmargin=*, noitemsep, topsep=0.2pt]
    \item We unify representative group-wise objectives through a common coefficient view and provide a controlled offline and online comparison, showing that the gains are robust across objectives.
    \item We identify two transferable design principles: training with groups of responses improves over pairwise training, and positive-response NLL regularization stabilizes group-wise optimization.
    \item We derive a sample-level surrogate for group-wise losses with matching first-order gradients.
    \item We show that this implementation reduces the peak memory overhead and enables training with larger groups while maintaining good efficiency.
\end{itemize}

\section{Related Work}
\vspace{-0.2em}
\textbf{Preference Alignment.}
Aligning LLMs with human preferences has relied on RLHF~\citep{christiano2017deep, ouyang2022training}, where a reward model trained on preference comparisons is optimized using policy gradient methods such as Proximal Policy Optimization (PPO)~\citep{schulman2017proximal}. 
While effective, RLHF introduces computational overhead and training instability due to extra reward-model training and reinforcement learning.

DPO~\citep{rafailov2023direct} provides a simpler alternative by reformulating preference alignment as a classification problem over chosen and rejected responses, allowing the policy to be optimized directly without training an explicit reward model. 
Building on this formulation, subsequent work has explored extensions to improve training stability and robustness. 
One direction augments the objective with supervised learning signals to mitigate issues such as likelihood degradation and over-optimization. 
For example, Regularized Preference Optimization (RPO)~\citep{liu2024provably} introduces a negative log-likelihood (NLL) term over chosen responses as an implicit regularizer, while Iterative Reasoning Preference Optimization (IRPO)~\citep{pang2024iterative} adopts a similar hybrid objective to stabilize online alignment for complex reasoning tasks. 
Another line of research investigates biases in preference learning, including length exploitation and overconfidence, and proposes calibration and normalization methods to address them~\citep{park2024disentangling, liu2024length, leng2025taming}.

Because DPO relies on a frozen reference model, several works have also explored reference-free formulations that simplify training. 
For example, SimPO~\citep{meng2024simpo} removes the reference model and constructs the preference objective using length-normalized log-likelihoods of the policy, while ORPO~\citep{hong2024orpo} integrates preference optimization directly into supervised fine-tuning through an odds-ratio objective.
Beyond human-labeled comparisons, recent self-improvement approaches such as SPIN~\citep{chen2024self} leverage iterative self-play to generate extra preference data for improvement during training.

\textbf{Group-wise Preference Learning.}
Although DPO operates on pairwise comparisons, modern preference datasets often contain multiple candidate responses for each prompt. 
Reducing these response sets to independent pairs discards useful relative information and limits the supervision signal available during training. 
To address this limitation, recent work has increasingly explored group-wise or listwise preference learning, drawing connections to classical learning-to-rank methods.

Several approaches extend preference optimization beyond pairwise comparisons using ranking-based objectives.
Preference Ranking Optimization (PRO)~\citep{song2024preference} and Listwise Preference Optimization (LiPO)~\citep{liu2025lipo} generalize pairwise comparisons to full rankings using the Plackett-Luce model~\citep{plackett1975analysis}.
Multi-Preference Optimization (MPO)~\citep{gupta2024multi} instead models preferences using a contrastive objective defined over the response groups.
Similar group-wise or listwise learning paradigms have also been explored in recommendation systems, where methods such as S-DPO~\citep{chen2024softmax}, LPO4Rec~\citep{li2025listwise}, and RankGR~\citep{fu2026rankgr} leverage large candidate sets to improve ranking and retrieval performance.

Despite these advances, prior work on group-wise preference optimization has two key limitations. First, algorithms are often proposed for different settings, lacking a unified comparison of group-wise objectives. 
Second, many methods operate on relatively small response groups due to the memory overhead of group-wise losses.
We address these limitations by systematically comparing representative methods under a unified experimental framework, studying larger group sizes, and introducing a memory-efficient surrogate implementation whose GPU memory usage does not scale with group size, enabling scalable training.

\section{Background}
\vspace{-0.2em}
\label{sec:background}

\textbf{Group DPO Setup.}
We consider a preference dataset where responses are organized into a collection $\mathcal G$ of groups.
Each group $g\in\mathcal G$ corresponds to a prompt $x$ and contains positive responses $P_g$ and negative responses $N_g$.
Responses within each group are treated as unordered, with no intra-set ranking. 
This setting arises naturally in practice: for example, in online training with rule-based outcome rewards (e.g., math reasoning), responses are often partitioned only by correctness, and in recommendation systems feedback typically indicates preferred vs.\ non-preferred items without ordering.

\textbf{Implicit preference score.}
Following DPO, we define the implicit preference score as follows:
\[
u_\theta(y|x) = \beta \big(\log \pi_\theta(y|x) - \log \pi_{\mathrm{ref}}(y|x)\big),
\]
a scaled log-likelihood ratio between the policy $ \pi_\theta$ and reference model $\pi_{\mathrm{ref}}$,
and $\beta$ is a scaling factor.

\textbf{Unified Group DPO Objective View.}
A broad class of preference objectives can be written as
\begin{equation}
\mathcal L_{\text{group}}(\theta)
=
\frac{1}{|\mathcal G|}\sum_{g\in\mathcal G}
\phi_g\left(u_{P_g}(\theta),u_{N_g}(\theta)\right)
\label{eq:group_obj}
\end{equation}
where $u_{P_g} = \{u_\theta(y|x) : y \in P_g \}$ and similarly for $u_{N_g}$,
and $\phi_g$ is method-specific, differing mainly in how scores are aggregated or contrasted within each group.
Despite differences in loss definitions, these objectives share a common computational issue: cross-response coupling within each group, which makes direct optimization memory-intensive. To address this, we introduce a memory-efficient and gradient-equivalent surrogate loss in Section~\ref{sec:method}.

\section{Memory-Efficient Surrogate}
\label{sec:method}
\label{sec:surrogate}

\textbf{Motivation: cross-response dependencies in group-wise objectives.}
Group-wise preference objectives couple responses within each prompt group. 
Many variants rely on cross-response interactions (e.g., score difference between positive and negative responses $u_p - u_n$ or softmax-based aggregations such as $\log \sum \exp(u_n)$), meaning that the gradient contribution of each response depends on the scores of other responses in the same group. 
As a result, the gradient for a single response cannot be computed from a purely local per-response loss. 

A straightforward implementation performs a joint forward pass over all responses in the group, computes the group-level loss, and backpropagates once through the resulting computation graph. This is analogous to the \emph{concatenated forward} strategy used in single-pair DPO, where chosen and rejected responses are processed together. In the group-wise setting, however, this approach requires retaining activations for \emph{all} responses. As a result, activation memory grows with the group size and becomes expensive for large group sizes or long sequences.

\textbf{Surrogate loss.}
To avoid constructing a joint computation graph over all responses, we formulate group-wise optimization as a sample-level surrogate objective with matched first-order gradients.

Let $\mathcal R=\bigcup_{g\in\mathcal G}(P_g\cup N_g)$ denote the response set in the current batch.
We first run a no-grad pass to obtain scores $\tilde u_i$, and use them to compute per-sample coefficients that quantify each response's contribution to the original group objective. Following Eq.~\eqref{eq:group_obj}, we define coefficients $c_i$, which are then treated as constants during backpropagation:
\vspace{-0.2em}
\[
c_i
=
\frac{1}{|\mathcal G|}
\left.
\frac{\partial \phi_g}{\partial u_i}
\right|_{u=\tilde u},
\quad i\in P_g\cup N_g.
\]
We then optimize the surrogate objective
\[
\mathcal L_{\text{sur}}(\theta)=\sum_{i\in\mathcal R} c_i\,u_i(\theta),
\]
which is first-order exact at the current parameter point: its gradient w.r.t.\ model parameters matches that of $\mathcal L_{\text{group}}$. Algorithm~\ref{alg:generic_precompute_coeff} summarizes the coefficient computation. 
Since the pass is no-grad, it does not retain any activations for backpropagation.
\begin{algorithm}[H]
\vspace{-0.2em}
\caption{Precompute Coeff for Group DPO}
\label{alg:generic_precompute_coeff}
\begin{algorithmic}[1]
\Require Group collection $\mathcal G$ with response partitions $(P_g,N_g)$; response set $\mathcal R=\bigcup_{g\in\mathcal G}(P_g\cup N_g)$; reference log-prob $r_i=\log \pi_{\mathrm{ref}}(y_i\mid x_i)$; scale $\beta$
\Ensure Coeff $\{c_i\}$ for all $i\in\mathcal R$

\State \textbf{No-grad pass:} $\tilde p_i \gets \log\pi_\theta(y_i\mid x_i),\ \forall i\in\mathcal R$
\State $\tilde u_i \gets \beta(\tilde p_i-r_i),\ \forall i\in\mathcal R$
\State Initialize $c_i\gets 0,\ \forall i\in\mathcal R$

\ForAll{$g\in\mathcal G$}
    \ForAll{$i\in P_g\cup N_g$}
        \State $c_i \gets c_i + \dfrac{1}{|\mathcal G|}\left.\dfrac{\partial \phi_g(u)}{\partial u_i}\right|_{u=\tilde u}$
    \EndFor
\EndFor
\State \Return $\{c_i\}_{i\in\mathcal R}$
\end{algorithmic}
\end{algorithm}
\vspace{-0.5em}
The choice of $\phi_g$ determines $c_i$.
Table~\ref{tab:dpo_coeff} summarizes the objectives and corresponding coefficients.

\textbf{Practical benefit.}
Because optimization operates on sample-level terms, backpropagation can be performed using standard token-level training primitives. 
As a result, peak activation memory becomes largely insensitive to group size (up to micro-batch effects; see Section~\ref{sec:efficiency_analysis} for more details), and sample-level techniques such as dynamic batch sizing and packing can be applied directly.

\begin{table*}[htbp]
\centering
\caption{Original per-group losses and derivatives.
Define $q(z)=\sigma(z)-1$.
We report $\partial \phi_g/\partial u_i$ for $i\in P_g$.}
\label{tab:dpo_coeff}
\small

\resizebox{\linewidth}{!}{%
\begin{tabularx}{\linewidth}{@{}l X l@{}}
\toprule
\textbf{Method} & \textbf{Original loss $\phi_g$} & \textbf{$\partial \phi_g/\partial u_i$} \\
\midrule

DPO 
& $\displaystyle -\log\sigma\big(u_i-u_j\big),\quad i\in P_g,\ j\in N_g$
& $\displaystyle q\big(u_i-u_j\big)$
\\

Margin
& $\displaystyle -\log\sigma\big(\bar u_{P_g}-\bar u_{N_g}\big)$
& $\displaystyle \frac{q\big(\bar u_{P_g}-\bar u_{N_g}\big)}{|P_g|}$
\\

MPO
& $\displaystyle -\log\left(\sum_{i\in P_g} s_i\right),\;\;
s_k=\frac{e^{u_k}}{\sum_{m\in P_g\cup N_g} e^{u_m}}$
& $\displaystyle s_i-\frac{s_i}{\rho},\;\; \rho=\sum_{i\in P_g} s_i$
\\

Softmax
& $\displaystyle \frac{1}{|P_g|}\sum_{i\in P_g}
-\log\sigma\Big(u_i-\log\sum_{j\in N_g}e^{u_j}\Big)$
& $\displaystyle \frac{q(z_i)}{|P_g|},\;\; z_i=u_i-\log\sum_{j\in N_g}e^{u_j}$
\\

All-Pairs
& $\displaystyle \frac{1}{|P_g||N_g|}\sum_{i\in P_g}\sum_{j\in N_g}
-\log\sigma\big(u_i-u_j\big)$
& $\displaystyle \frac{1}{|P_g||N_g|}\sum_{j\in N_g}
q\big(u_i-u_j\big)$
\\

\bottomrule
\end{tabularx}%
}
\vspace{-1.0em}
\end{table*}

\section{Experiments}
\label{sec:experiments}
In this section, we conduct a series of experiments using \texttt{gemma3-4b-sft}\footnote{A SFT checkpoint from \texttt{gemma-3-4b-pt} and trained on \href{https://huggingface.co/datasets/allenai/Dolci-Instruct-SFT-No-Tools}{allenai/Dolci-Instruct-SFT-No-Tools}~\citep{olmo2025olmo}.}~\citep{gemmateam2025gemma3technicalreport}, together with \texttt{olmo-3-7b-it-sft} and \texttt{olmo-3.1-32b-it-sft}~\citep{olmo2025olmo}, in the \textbf{offline} setting, and \texttt{qwen3-4b-base}~\citep{yang2025qwen3} in the \textbf{online} setting, to address the following questions: 
(1) Is Group DPO more beneficial than pairwise comparisons?
(2) Is the negative log-likelihood (NLL) loss term necessary in the group-wise setting?
(3) What are the practical memory and latency benefits of our surrogate implementation?
Details of data preparation, evaluation, and hyperparameters are provided in Appendix~\ref{appendix:experimental_details}.

\textbf{Methods.}
We compare several group-wise preference objectives, including Margin, MPO, All-Pairs, and Softmax (see Table~\ref{tab:dpo_coeff} for their loss expressions), along with vanilla DPO as a baseline.
\begin{itemize}[leftmargin=*, itemsep=0pt, parsep=0pt, topsep=0pt]
    \item \textbf{RFT}. Rejection Fine-Tuning, which trains only on positive responses within each prompt group.

    \item \textbf{DPO}~\citep{rafailov2023direct}. Direct Preference Optimization, which only trains on single pairs\footnote{Unless otherwise specified, DPO in experiments uses randomly selected pairs.}.

    \item {\textbf{Margin}}. A group-level variant optimizing mean reward gap between positive and negative groups.

    \item {\textbf{MPO}}~\citep{gupta2024multi}. Multi-Preference Optimization maximizing probability over positives.

    \item {\textbf{All-Pairs}}. A variant averaging the DPO loss over all positive-negative pairs in each group. Explicitly materializing these gives the pair-expansion implementation denoted as \emph{flatten} in Section~\ref{sec:efficiency_analysis}.

    \item {\textbf{Softmax}}~\citep{chen2024softmax}. The loss contrasts one positive against softmax-aggregated negatives; extended by aggregating over all positives.
\end{itemize}
We implement all methods using our proposed surrogate implementation and denote them with a ``+'' suffix in the tables. 
Importantly, the ``+'' suffix indicates a change in implementation rather than in the underlying objective: the corresponding ``+'' and vanilla variants use the same response groups and optimize the same objective, differing only in how backpropagation is executed. Appendix~\ref{appendix:surrogate_sanity} verifies their gradient and multi-step update agreement across objectives, optimizers, learning rates, and group sizes. The vanilla implementations of the group-wise objectives exceed GPU memory limits under the main experimental setting, whereas the surrogate implementations enable training in this regime (see Section~\ref{sec:efficiency_analysis}).
Unless otherwise specified, all methods include an NLL loss term with coefficient 1.0, which is commonly used in preference optimization to prevent the log probability of positive responses from decreasing~\citep{liu2024provably, pang2024iterative, wang2024preference, grattafiori2024llama, pal2024smaug}. 
 We further apply a length-normalized loss to mitigate length bias~\citep{olmo2025olmo, lambert2024tulu, meng2024simpo, yuan2023rrhf}, as shown below:
\[
\begin{aligned}
\min_{\pi_\theta}\ 
\mathbb{E}_{(x, y_p, y_n)\sim \mathcal{D}}
\Bigg[
&-\log \sigma\Bigg(
\frac{\beta}{|y_p|}
\log \frac{\pi_\theta(y_p \mid x)}{\pi_{\mathrm{ref}}(y_p \mid x)}
\\
&\qquad
-
\frac{\beta}{|y_n|}
\log \frac{\pi_\theta(y_n \mid x)}{\pi_{\mathrm{ref}}(y_n \mid x)}
\Bigg)
\\
&\qquad
-\alpha\,\frac{1}{|y_p|}\log \pi_\theta(y_p \mid x)
\Bigg]
\end{aligned}
\]
\textbf{Offline Setting.}
We use prompts sourced from \href{https://huggingface.co/datasets/allenai/Dolci-Instruct-DPO}{Dolci-Instruct-DPO}~\citep{olmo2025olmo}.
For each prompt, candidate responses are sampled from multiple models and evaluated using a reward model.\footnote{ \href{https://huggingface.co/Skywork/Skywork-Reward-V2-Qwen3-8B}{Skywork-Reward-V2-Qwen3-8B}~\citep{liu2025skywork}.}
The top-$k$ responses form the positive group, while the bottom-$k$ form the negative group.
Importantly, the reward scores are discarded after grouping, and responses within each group are treated as unordered; that is, no intra-group ranking is assumed.

For the 4B and 7B models, we use a group size of 16 (8 positive and 8 negative responses) and sweep $\beta \in \{2.0, 5.0, 10.0, 15.0\}$ with a fixed learning rate of $1e\!-\!6$. 
For the 32B model, the group size is reduced to 8 and we fix $\beta=2.0$ due to computational constraints. 
Additional training details and full results are provided in Appendices \ref{appendix:experimental_details} and~\ref{appendix:full_results}.

\textbf{Online Setting.}
We use prompts sourced from \href{https://huggingface.co/datasets/BytedTsinghua-SIA/DAPO-Math-17k}{DAPO-Math-17k}~\citep{yu2026dapo}.
For each prompt, multiple responses are sampled from the policy model. 
A rule-based reward function is used to evaluate response accuracy, and the responses are partitioned into positive and negative groups based on correctness of the final answer.
Unlike the offline setting, the resulting groups may be \emph{\textbf{unbalanced}}; we only enforce the total group size, while the number of positive and negative responses can vary.
During training, we use a learning rate of $1e\!-\!6$ and set $\beta=5.0$ for all models and methods.


\subsection{Q1: Group-wise Optimization Consistently Beats Pairwise Baselines} \label{sec:groupisbetter}
\begin{table*}[ht]
\centering
\scriptsize
\setlength{\tabcolsep}{1.5pt}
\renewcommand{\arraystretch}{1.10}
\resizebox{\linewidth}{!}{%
\begin{tabularx}{\linewidth}{@{}>{\raggedright\arraybackslash}p{0.11\linewidth}ccccccc@{\hspace{1.5pt}}!{\color{tableblue}\vrule width 1.4pt}@{\hspace{1.5pt}}ccccccc@{\hspace{1.5pt}}!{\color{tableblue}\vrule width 1.4pt}@{\hspace{1.5pt}}ccccccc@{}}
\toprule
 & \multicolumn{7}{c}{gemma-3-4b-sft} & \multicolumn{7}{c}{olmo-3-7b-it-sft} & \multicolumn{7}{c}{olmo-3.1-32b-it-sft} \\
\cmidrule(lr){2-8} \cmidrule(lr){9-15} \cmidrule(lr){16-22}
 & \makebox[2.3em][c]{Base} & \makebox[2.3em][c]{RFT} & \makebox[2.3em][c]{DPO} & \makebox[2.3em][c]{Mag+} & \makebox[2.3em][c]{MPO+} & \makebox[2.3em][c]{Pair+} & \makebox[2.3em][c]{Soft+} & \makebox[2.3em][c]{Base} & \makebox[2.3em][c]{RFT} & \makebox[2.3em][c]{DPO} & \makebox[2.3em][c]{Mag+} & \makebox[2.3em][c]{MPO+} & \makebox[2.3em][c]{Pair+} & \makebox[2.3em][c]{Soft+} & \makebox[2.3em][c]{Base} & \makebox[2.3em][c]{RFT} & \makebox[2.3em][c]{DPO} & \makebox[2.3em][c]{Mag+} & \makebox[2.3em][c]{MPO+} & \makebox[2.3em][c]{Pair+} & \makebox[2.3em][c]{Soft+} \\
\midrule
\addlinespace[2pt]
\rowcolor{tablepink}\multicolumn{22}{@{}l}{\textbf{Math}}\\
\addlinespace[2pt]
\quad AIME24 & - & - & - & - & - & - & - & 7.5 & 15.8 & 25.2 & 29.6 & 32.1 & 32.1 & 31.9 & 14.2 & 27.1 & 32.1 & 40.8 & 46.0 & 46.7 & 44.2 \\
\quad AIME25 & - & - & - & - & - & - & - & 7.9 & 17.3 & 25.2 & 28.1 & 26.0 & 26.9 & 26.5 & 8.1 & 27.5 & 25.6 & 33.3 & 35.8 & 31.7 & 32.3 \\
\quad AMC23 & 15.0 & 14.7 & 16.2 & 16.7 & 16.6 & 15.6 & 14.5 & 42.0 & 54.7 & 61.6 & 67.8 & 73.1 & 69.5 & 72.8 & 52.3 & 68.0 & 67.7 & 76.6 & 78.3 & 77.2 & 77.5 \\
\quad MATH500 & 35.6 & 38.4 & 40.2 & 40.4 & 39.6 & 40.0 & 40.2 & 69.4 & 77.6 & 81.0 & 86.6 & 88.4 & 89.2 & 87.8 & 76.4 & 83.0 & 86.6 & 90.6 & 90.0 & 90.8 & 89.8 \\
\quad Minerva & 17.6 & 16.9 & 18.4 & 17.3 & 18.4 & 20.2 & 21.3 & 30.5 & 33.1 & 34.6 & 38.2 & 34.9 & 38.2 & 39.3 & 39.3 & 43.0 & 45.2 & 43.8 & 43.8 & 44.5 & 43.4 \\
\quad Olympiad & 9.6 & 11.0 & 10.8 & 11.9 & 13.2 & 13.5 & 10.8 & 34.5 & 43.7 & 51.4 & 56.0 & 57.3 & 57.3 & 57.3 & 43.7 & 52.9 & 59.9 & 61.3 & 64.7 & 67.0 & 64.1 \\
\textbf{Math Avg} & 19.5 & 20.2 & 21.4 & 21.6 & 21.9 & \textbf{\textit{22.3}} & 21.7 & 32.0 & 40.4 & 46.5 & 51.1 & 52.0 & 52.2 & \textbf{\textit{52.6}} & 39.0 & 50.2 & 52.8 & 57.7 & \textbf{\textit{59.8}} & 59.6 & 58.5 \\
\addlinespace[2pt]
\rowcolor{tablepink}\multicolumn{22}{@{}l}{\textbf{Reasoning}}\\
\addlinespace[2pt]
\quad AGIEval Eng & 49.4 & 49.0 & 50.3 & 50.5 & 51.2 & 51.2 & 51.2 & 54.1 & 55.8 & 58.8 & 59.4 & 60.9 & 61.8 & 61.4 & 67.7 & 69.0 & 73.6 & 72.6 & 74.3 & 74.4 & 74.4 \\
\quad GPQA-D & 22.3 & 21.4 & 26.3 & 24.1 & 25.5 & 23.6 & 26.7 & 32.1 & 27.9 & 33.7 & 36.9 & 37.3 & 36.8 & 35.4 & 34.1 & 33.2 & 47.8 & 46.6 & 48.5 & 48.9 & 49.1 \\
\quad MMLU-PRO & 31.5 & 32.8 & 36.7 & 34.7 & 36.2 & 33.5 & 37.5 & 45.6 & 46.3 & 51.6 & 55.0 & 54.8 & 55.2 & 53.6 & 57.0 & 58.2 & 67.9 & 67.3 & 69.3 & 69.4 & 69.1 \\
\quad ZebraLogic & 15.6 & 11.7 & 18.9 & 16.9 & 16.8 & 16.9 & 18.3 & 17.5 & 15.8 & 25.4 & 26.4 & 28.8 & 30.9 & 25.5 & 30.0 & 28.3 & 19.1 & 12.1 & 18.2 & 21.5 & 18.6 \\
\quad BBH & 40.4 & 49.9 & 50.5 & 49.3 & 51.7 & 50.0 & 51.4 & 46.0 & 66.7 & 71.5 & 73.2 & 73.6 & 73.8 & 72.9 & 60.0 & 77.6 & 81.6 & 81.6 & 82.3 & 82.1 & 81.6 \\
\addlinespace[2pt]
\rowcolor{tablepink}\multicolumn{22}{@{}l}{\textbf{Coding}}\\
\addlinespace[2pt]
\quad HumanEval+ & 32.8 & 56.8 & 57.5 & 58.0 & 52.3 & 48.5 & 58.5 & 73.9 & 72.0 & 74.5 & 73.4 & 73.6 & 73.2 & 65.9 & 85.3 & 83.5 & 88.5 & 84.8 & 85.3 & 87.6 & 84.5 \\
\quad LiveCode v6 & 9.4 & 10.7 & 10.6 & 11.5 & 10.7 & 10.2 & 9.8 & 14.9 & 18.7 & 19.5 & 22.5 & 19.5 & 18.0 & 16.1 & 21.4 & 25.3 & 29.0 & 28.6 & 26.6 & 29.7 & 27.5 \\
\addlinespace[2pt]
\rowcolor{tablepink}\multicolumn{22}{@{}l}{\textbf{Instruction Following}}\\
\addlinespace[2pt]
\quad IFBench & 20.4 & 16.0 & 22.8 & 21.1 & 21.1 & 24.5 & 21.8 & 28.6 & 21.1 & 29.6 & 30.3 & 31.3 & 34.4 & 32.7 & 32.7 & 24.1 & 36.1 & 35.4 & 33.3 & 35.0 & 34.4 \\
\quad IFEval & 62.5 & 58.4 & 66.7 & 63.6 & 66.9 & 65.6 & 64.9 & 80.6 & 73.4 & 83.4 & 84.3 & 84.1 & 82.1 & 83.9 & 87.6 & 83.0 & 87.1 & 86.7 & 85.8 & 87.2 & 85.6 \\
\textbf{General Avg} & 31.6 & 34.1 & \textbf{\textit{37.8}} & 36.6 & 36.9 & 36.0 & 37.8 & 43.7 & 44.2 & 49.8 & 51.3 & 51.6 & \textbf{\textit{51.8}} & 49.7 & 52.9 & 53.6 & 59.0 & 57.3 & 58.2 & \textbf{\textit{59.5}} & 58.3 \\
\midrule
\textbf{Avg} & 27.9 & 29.8 & 32.8 & 32.0 & 32.3 & 31.8 & \textbf{\textit{32.8}} & 39.0 & 42.7 & 48.5 & 51.2 & 51.7 & \textbf{\textit{52.0}} & 50.9 & 47.3 & 52.2 & 56.5 & 57.5 & 58.8 & \textbf{\textit{59.6}} & 58.4 \\
\bottomrule
\end{tabularx}%
}
\caption{Performance comparison of different methods across multiple models. Results are reported on math, reasoning, coding, and instruction-following benchmarks. For each method, we report the checkpoint with the highest validation average. ``General Avg'' denotes the average over non-math tasks, and ``Avg'' over all benchmarks.}\label{tab:main_table}
\end{table*}
\textbf{Offline Setting.}
The results for the offline setting are presented in Table~\ref{tab:main_table}\footnote{We exclude \texttt{gemma3-4b} from the AIME eval because its performance is too low to yield meaningful results.}.
During training, we evaluate the model on math tasks every 100 steps.
For the final evaluation, we compare the checkpoint with the highest validation performance on math tasks and the last checkpoint, and report the better of the two.
As the results show, group-wise variants consistently outperform single-pair DPO and RFT.

Although our experimental setting assumes no intra-group ranking, for a fairer comparison we also report results for single-pair DPO constructed from the best positive sample and the worst negative sample (i.e., those with the highest and lowest reward scores).
This construction provides a stronger pairwise baseline by selecting the most separable pair from each group, which is commonly used when building binarized DPO datasets~\citep{cui2023ultrafeedback}.
The corresponding results are reported in Appendix~\ref{appendix:full_results}, Table~\ref{tab:dpo-edge}.
Although the performance gap becomes smaller under this setting, the overall observation and conclusion remain unchanged.

These results indicate that incorporating multiple responses within each prompt group is beneficial.
Moreover, we observe that the performance of different group-wise variants is largely comparable.

\textbf{Online Setting.}
\begin{table*}[ht]
\centering
\scriptsize
\setlength{\tabcolsep}{1.8pt}
\renewcommand{\arraystretch}{1.05}
\resizebox{\linewidth}{!}{%
\begin{tabularx}{\linewidth}{@{}>{\raggedright\arraybackslash}Xccccccccccc@{}}
\toprule
\multicolumn{12}{c}{qwen3-4b-base} \\
\cmidrule(lr){2-12}
\multirow{2}{*}{Dataset} & \multicolumn{1}{c}{RFT} & \multicolumn{2}{c}{DPO} & \multicolumn{2}{c}{All-Pairs+} & \multicolumn{2}{c}{Margin+} & \multicolumn{2}{c}{MPO+} & \multicolumn{2}{c}{Softmax+} \\
\cmidrule(lr){2-2} \cmidrule(lr){3-4} \cmidrule(lr){5-6} \cmidrule(lr){7-8} \cmidrule(lr){9-10} \cmidrule(lr){11-12}
 & nll=0.0 & nll=0.0 & nll=1.0 & nll=0.0 & nll=1.0 & nll=0.0 & nll=1.0 & nll=0.0 & nll=1.0 & nll=0.0 & nll=1.0 \\
\midrule
\addlinespace[2pt]
\rowcolor{tablepink}\multicolumn{12}{@{}l}{\textbf{Math}}\\
\addlinespace[2pt]
\quad AIME24 & 16.2 & 17.5 & 18.1 & 20.4 & 19.4 & 22.5 & 22.5 & 20.0 & 21.7 & 18.3 & 18.8 \\
\quad AIME25 & 12.9 & 14.2 & 18.1 & 15.8 & 19.8 & 19.0 & 19.4 & 16.2 & 18.3 & 16.9 & 17.7 \\
\quad AMC23 & 61.7 & 59.4 & 61.1 & 60.8 & 66.7 & 62.7 & 68.0 & 62.3 & 65.3 & 62.7 & 67.0 \\
\quad MATH500 & 82.0 & 83.0 & 85.4 & 85.0 & 83.4 & 85.2 & 84.4 & 83.4 & 84.6 & 81.6 & 86.2 \\
\quad Minerva & 39.0 & 38.6 & 42.3 & 42.3 & 38.2 & 40.1 & 43.8 & 41.2 & 44.1 & 37.9 & 43.4 \\
\quad Olympiad & 46.7 & 48.1 & 49.0 & 51.0 & 51.9 & 50.2 & 47.9 & 50.2 & 51.9 & 47.4 & 50.8 \\
\textbf{Math Avg} & 43.1 & 43.5 & 45.7 & 45.9 & 46.6 & 46.6 & 47.6 & 45.6 & \textbf{\textit{47.6}} & 44.1 & 47.3 \\
\bottomrule
\end{tabularx}%
}
\caption{Performance comparison of different methods for \texttt{qwen3-4b-base}. We report the highest checkpoint.}
\label{tab:online_table}
\end{table*}

The overall results for the online setting are presented in Table~\ref{tab:online_table}. 
During training, we evaluate the model on math tasks every 5 rollout steps and report the checkpoint with the highest performance.
Consistent with the offline setting, group-wise variants outperform the RFT and pairwise DPO baseline.
Among the group-wise variants, the performance differences are relatively small, with All-Pairs, Margin, MPO, and Softmax all achieving comparable average performance.

\textbf{Effect of Group Size.}
\begin{figure}[ht]
\centering
\includegraphics[width=0.5\textwidth]{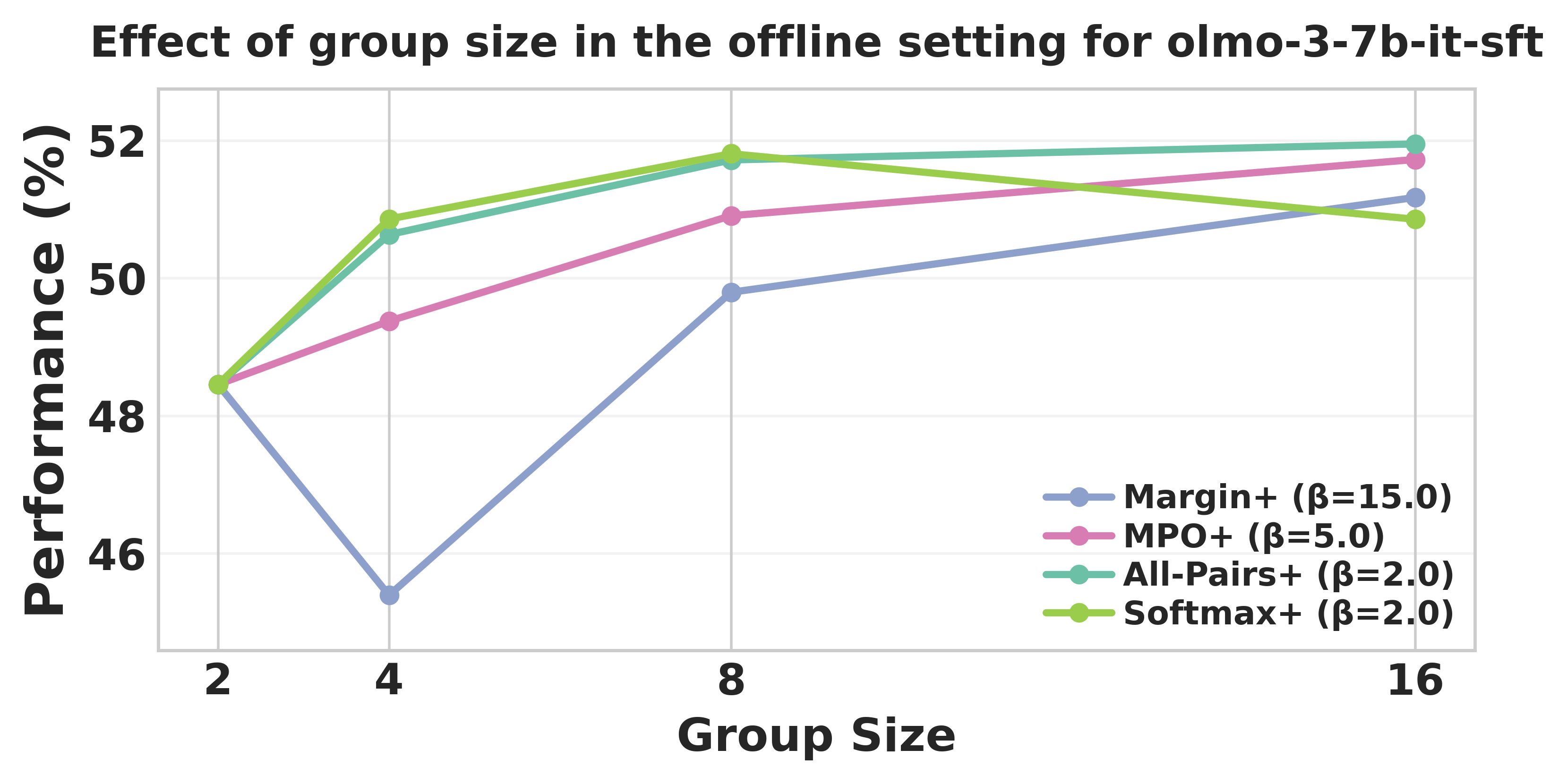}
\caption{Effect of group size for \texttt{olmo-3-7b-it-sft}.}
\vspace{-0.8em}
\label{fig:group_size_effect}
\end{figure}
Figure~\ref{fig:group_size_effect} demonstrates the impact of group size on performance in the offline setting using \texttt{olmo-3-7b-it-sft}.
As the group size increases from 2 to 8, performance consistently improves across most group-wise variants, suggesting that incorporating more responses per prompt provides richer and more informative comparative signal during training.
However, gains begin to plateau beyond this range, with only marginal improvements or slight decreases observed at group size 16, likely due to diminishing returns from additional comparisons and increased noise.
Overall, these results indicate that moderate group sizes are sufficient to capture most of the benefits of group-wise comparisons, while larger groups provide limited additional advantage. 
Consistent with our earlier observations, the performance differences across group-wise variants remain small.

\textbf{Why do groups help?}
For the All-Pairs objective, the benefit of using the full response group has a direct variance-reduction interpretation.
Consider a random-pair DPO baseline that samples $I$ uniformly from $P_g$ and $J$ uniformly from $N_g$ for each prompt group, and let
$g_{\mathrm{pair}}=\nabla_\theta \ell_{IJ}(\theta)$ denote its stochastic gradient, where $\ell_{ij}=-\log\sigma(u_i-u_j)$.
All-Pairs gradient averages the loss over all pairs:
\[
\begin{aligned}
g_{\mathrm{AP}}
&=\frac{1}{|P_g||N_g|}\sum_{i\in P_g}\sum_{j\in N_g}
\nabla_\theta \ell_{ij}(\theta) \\
&=\mathbb E[g_{\mathrm{pair}}\mid x,P_g,N_g].
\end{aligned}
\]
Thus, All-Pairs is the Rao--Blackwellized version of random-pair DPO conditioned on the observed response group.
Letting $\mu=\mathbb E[g_{\mathrm{pair}}]=\mathbb E[g_{\mathrm{AP}}]$, the conditional-variance decomposition gives
\[
\mathbb E\|g_{\mathrm{pair}}-\mu\|^2
=\mathbb E\|g_{\mathrm{AP}}-\mu\|^2
+\mathbb E\|g_{\mathrm{pair}}-g_{\mathrm{AP}}\|^2.
\]
Since the final term is nonnegative,
$\mathbb E\|g_{\mathrm{AP}}-\mu\|^2\leq\mathbb E\|g_{\mathrm{pair}}-\mu\|^2$.
Thus, All-Pairs preserves the expected gradient of the prompt-level pairwise objective while reducing the variance caused by selecting an arbitrary response pair.
This variance-reduction view complements the analysis of MPO~\citep{gupta2024multi}, which bounds its expected alignment/attribute bias by $\widetilde{\mathcal O}(C/\sqrt{n})$, where $n$ is the number of preferred responses per prompt and $C$ depends on the maximum attribute variance.

Table~\ref{tab:dpo_coeff} provides further intuition for these objectives.
All-Pairs weights each response according to its average violation against the opposite set; Softmax uses its log-sum-exp aggregation to emphasize high-scoring negatives, acting as smooth hard-negative mining; and MPO contrasts the probability mass assigned to the positive set against the full response set.
These objectives therefore use the full group to shape prompt-level score separation, rather than relying on a single sampled comparison.
Consistently, Figure~\ref{fig:training_dynamics_diagnostics} shows that All-Pairs has lower and more stable gradient norms than random-pair DPO, while all group-wise objectives achieve larger positive-negative implicit reward margins.


\subsection{Q2: Positive-Sample NLL is Critical for Training Stability}
\label{sec:nll_stability}
\begin{figure}[ht]
  \centering
\vspace{-0.5em}
\includegraphics[width=\columnwidth]{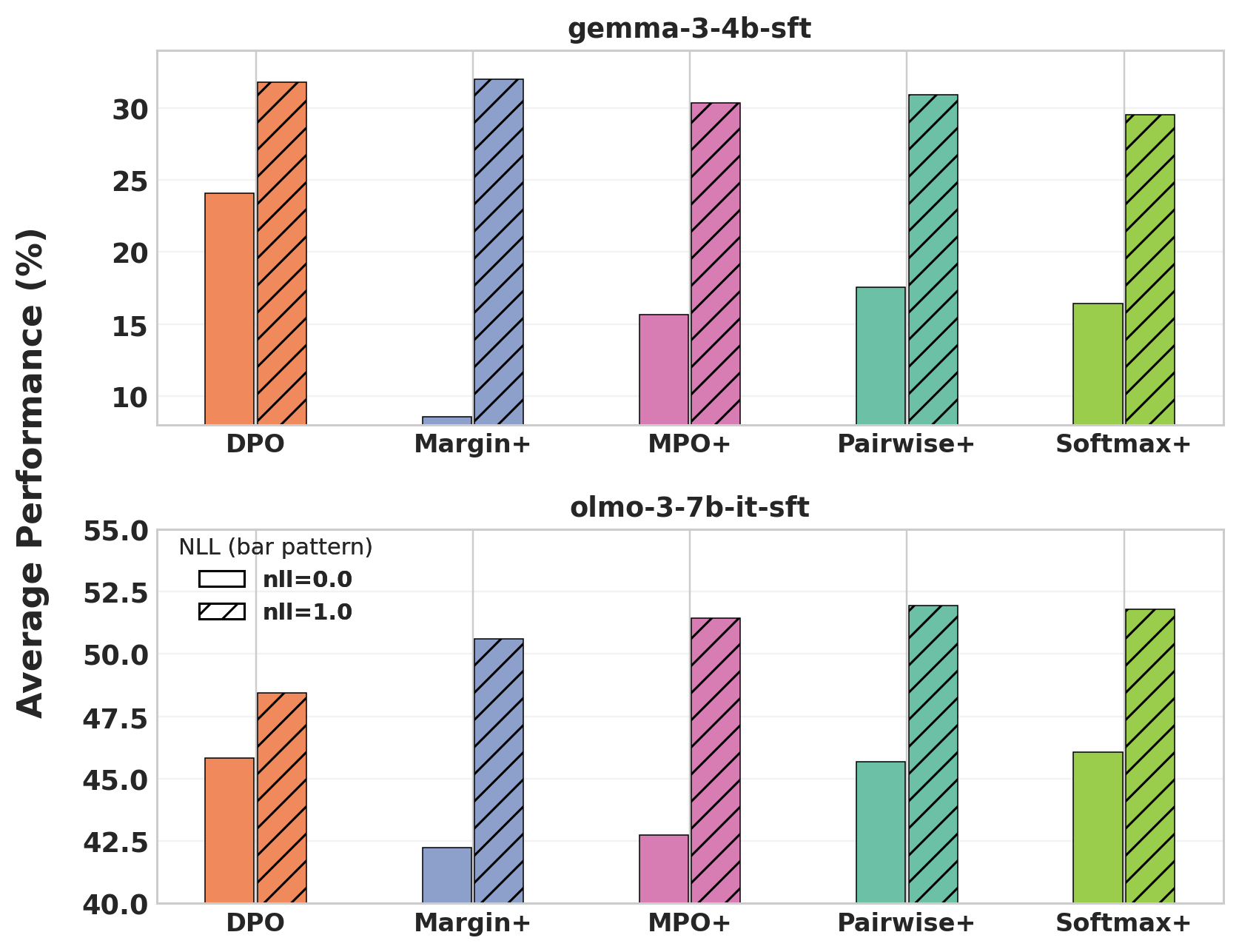}
\caption{Offline results with and without the NLL term ($\alpha=1.0$ vs.\ $\alpha=0.0$). Bars show ``General Avg.''; including NLL consistently improves performance.}
\label{fig:offline_nll}
\end{figure}
\begin{figure}[ht]
\centering
\includegraphics[width=\columnwidth]{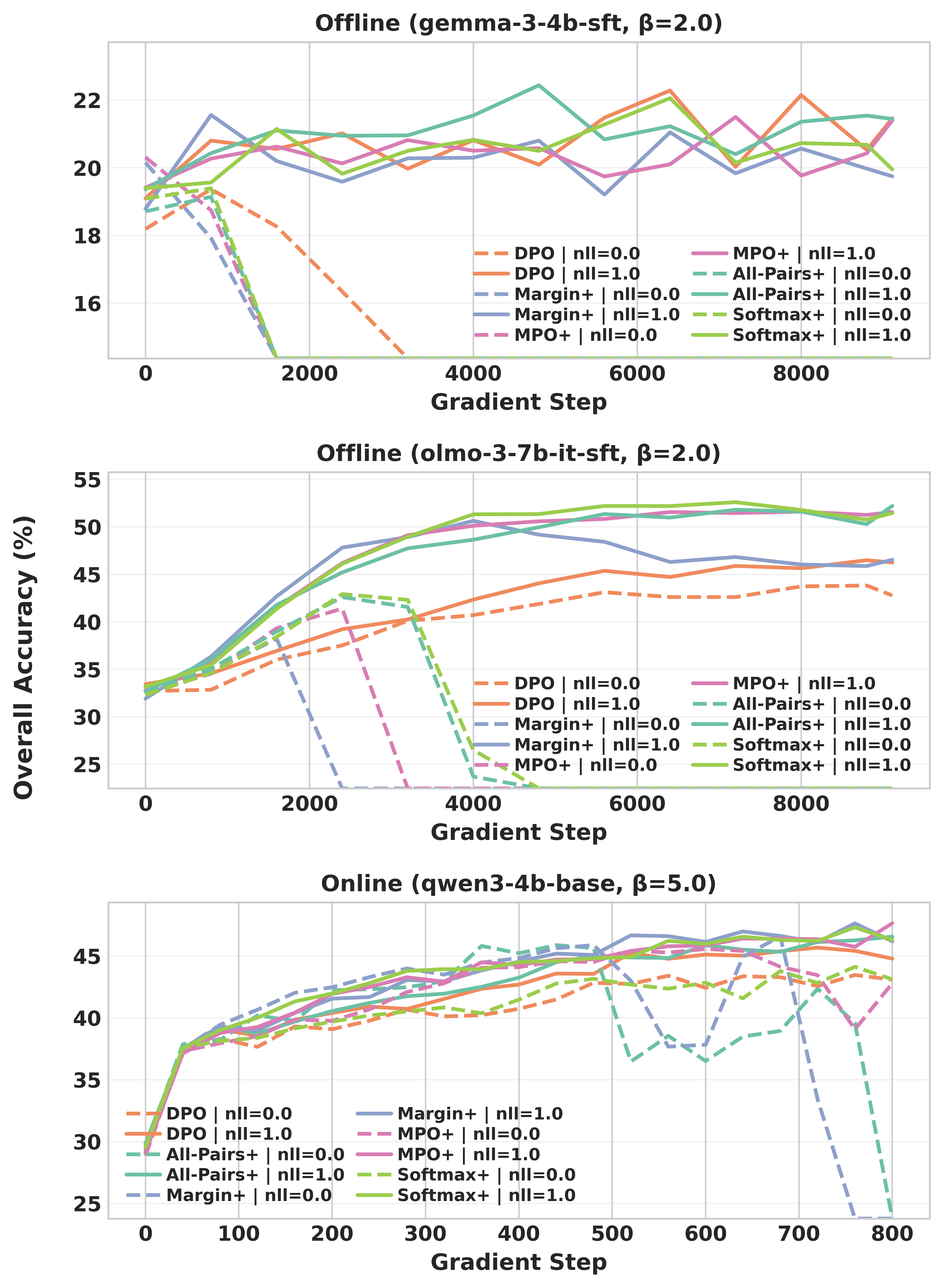}
\caption{Training curves in the offline (top two figures) and online (bottom figure) settings for three models.}
\vspace{-1.0em}
\label{fig:training_dynamic}
\end{figure}
\textbf{NLL term stabilizes training and improves performance.}
Figure~\ref{fig:offline_nll} compares average performance with and without the NLL term.
Across methods and both evaluated models, including the NLL term consistently improves performance,
indicating clear benefits in the offline setting.
A similar trend appears in the online setting: as shown in Table~\ref{tab:online_table}, removing the term consistently results in lower final accuracy.
Moreover, training without the NLL term is often unstable. Runs with NLL improve steadily, whereas several methods without it collapse after initial gains, as shown in Figure~\ref{fig:training_dynamic}.

Overall, these results suggest that the NLL term acts as an important regularizer in the group-wise setting, improving stability and preventing degradation of log probabilities on positive responses.
Appendix~\ref{appendix:training_dynamics} provides a direct diagnostic: without NLL, the chosen implicit reward decreases substantially for all four group-wise objectives, whereas adding NLL largely prevents this degradation.

\subsection{Q3: What are the memory and latency benefits of our surrogate implementation?}\label{sec:efficiency_analysis}
The empirical gains observed in Section~\ref{sec:groupisbetter} rely heavily on the ability to scale group sizes without exhausting GPU memory. Here, we quantify the hardware efficiency of our surrogate approach, demonstrating how it avoids the memory bottlenecks of standard group-wise implementations.

\textbf{Analysis Setups.}
Following the efficiency analysis~\citep{yang2024s}, we report two training-efficiency metrics:
(1) \textbf{memory overhead}, as
\[
\begin{aligned}
\mathrm{overhead\_mem}
&=\max_{\substack{\mathrm{forward}\\+\mathrm{backward}}}
\!\left(\mathrm{mem\_allocated}\right)
\\
&\quad-\mathrm{base\_mem}
\end{aligned}
\]
where \(\mathrm{base\_mem}\) is measured after warmup full steps, once optimizer states have been initialized; thus it includes model parameters and optimizer states.
This metric captures the additional memory footprint during training from activations and gradients, while excluding temporary allocations from \texttt{optimizer.step()}.
(2) \textbf{step latency}, measured as the average wall-clock time of a full training step (forward + backward + optimizer step) after warmup.
For all measurements, we use 3 full warmup steps before memory measurement, and 2 warmup steps with 5 measured steps for latency.
We enable gradient checkpointing in all reported results to reflect practical model training settings.

All analyses are conducted on a single NVIDIA H100 (80GB) SXM GPU.
We compare three implementations of the All-Pairs objective:
(i) \textbf{\emph{vanilla}}, which processes all responses in a group in a single forward-backward pass;
(ii) \textbf{\emph{flatten}}, which pre-converts each group into pairs and processes them sequentially to avoid a joint computation graph; and
(iii) \textbf{\emph{surrogate}}, our surrogate implementation.
These are three implementations of the same learning objective rather than three different objectives.
Vanilla retains one group-coupled graph, flatten explicitly materializes and backpropagates the $|P_g||N_g|$ DPO terms, and surrogate first captures the same cross-response interactions in group-dependent coefficients before performing response-level backpropagation.
Because our binary-group setting compares only positive-negative responses, pair expansion produces $|P_g||N_g|$ pairs---not all $G(G-1)/2$ unordered response pairs; for balanced groups, $|P_g|=|N_g|=G/2$ and the count is $G^2/4$.
The implementations also have different natural backward micro-batch units.
After coefficient precomputation, the surrogate can backpropagate one response at a time, so its minimum unit is one response; flatten must jointly evaluate a chosen-rejected pair, so its minimum unit is one pair, or two responses; and vanilla retains all responses in the group-coupled graph.
Our measurements use these natural minimum units while holding the underlying groups and All-Pairs objective fixed.
\begin{table}[ht]
\centering
\small
\vspace{-0.5em}
\setlength{\tabcolsep}{6pt}
\begin{tabularx}{\linewidth}{X r}
\toprule
Method & Asymptotic Complexity \\
\midrule
Vanilla   & $\mathcal{O}(G\,C(T) + G^2)$ \\
Flatten   & $\mathcal{O}(G^2\,C(T))$ \\
Surrogate & $\mathcal{O}(G\,C(T) + G^2)$ \\
\bottomrule
\end{tabularx}
\caption{
Per-step asymptotic time complexity of the three implementations in our efficiency benchmark,
under the balanced-group setting used in our experiments, where each group contains
$G$ responses with an equal split, $|P_g|=|N_g|=G/2$.
Here $C(T)$ denotes the cost of one sequence-level LLM forward/backward computation
for sequence length $T$.
$G^2$ term comes from group-level interactions across all positive-negative pairs.
}
\vspace{-0.2em}
\label{tab:efficiency_complexity}
\end{table}

\textbf{Results.}
\begin{figure}[ht]
  \centering
  \begin{subfigure}{\columnwidth}
    \centering
    \includegraphics[width=\columnwidth]{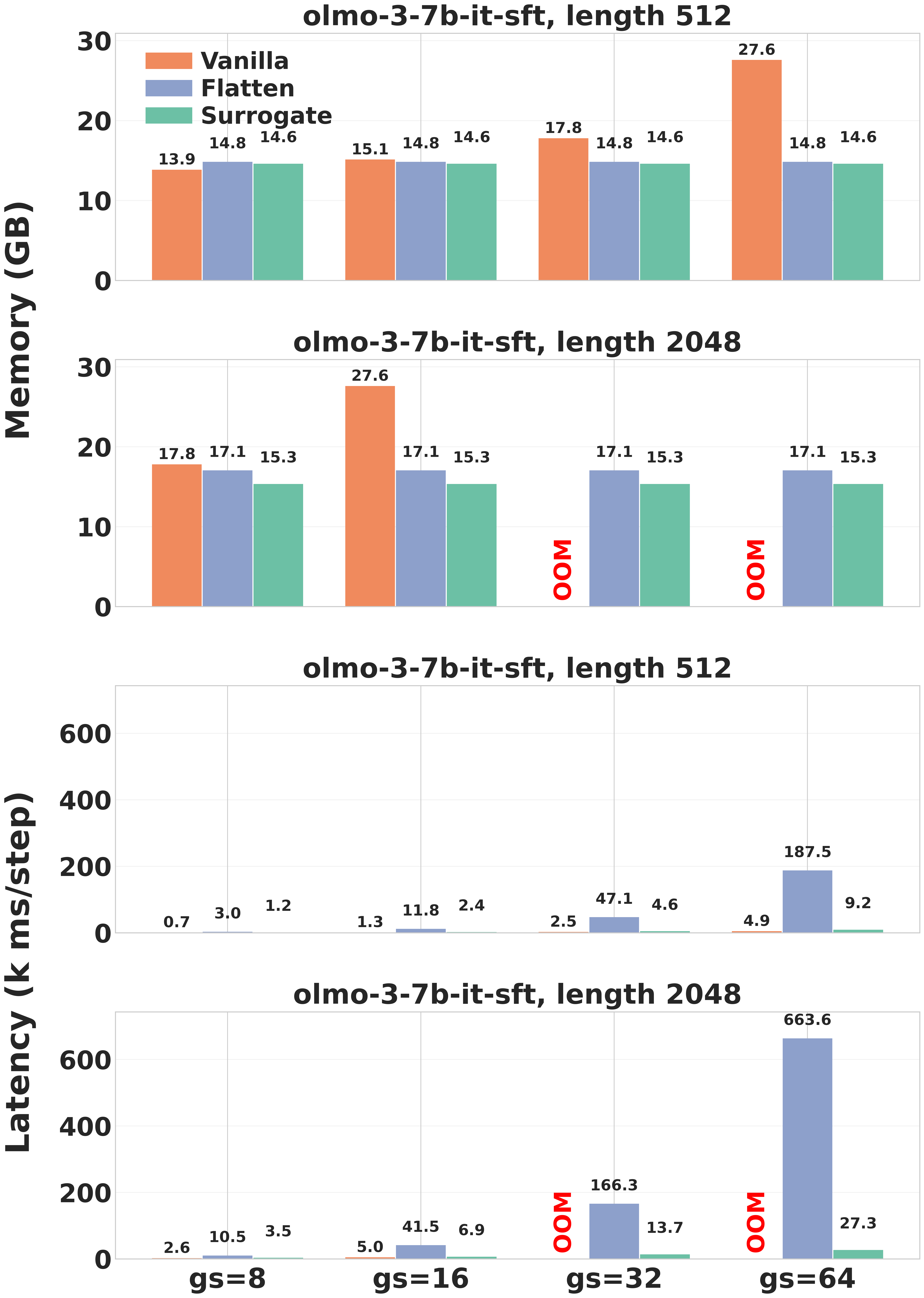}
  \end{subfigure}
  \caption{Peak memory overhead and average step latency for \texttt{olmo-3-7b-it-sft} across sequence lengths and group sizes. The surrogate reduces memory vs.\ the vanilla implementation with modest latency overhead.}
  \vspace{-1.3em}
  \label{fig:efficiency}
\end{figure}
Table~\ref{tab:efficiency_complexity} provides the theoretical picture, and Figure~\ref{fig:efficiency} confirms the same trend empirically.

\emph{\textbf{Vanilla}} becomes increasingly memory-intensive as the group size grows.
Because all responses in a group are coupled in a single forward-backward computation graph, activation memory increases rapidly with group size, often leading to OOM errors at larger groups or longer sequence lengths.

\emph{\textbf{Flatten}} exhibits the opposite trade-off.
By avoiding a joint group graph and processing $P$-$N$ pairs sequentially, it keeps memory relatively stable, but incurs substantial latency overhead.
Furthermore, its explicit pair expansion makes the expensive model computation scale quadratically with group size in the balanced setting, limiting scalability.

In contrast, \emph{\textbf{surrogate}} achieves a substantially better practical trade-off.
It maintains low and nearly group-size-stable memory usage, while remaining much faster than \emph{\textbf{flatten}}.
Although \emph{\textbf{surrogate}} introduces an additional no-grad coefficient precomputation pass, it avoids the quadratic growth in LLM forward/backward calls required by \emph{\textbf{flatten}}.
Its remaining quadratic term comes only from lightweight group-level pair interactions, rather than repeated model evaluations.
As a result, surrogate matches the asymptotic order of \emph{\textbf{vanilla}} while preserving a much more favorable memory profile.

\section{Conclusion}
\vspace{-0.2em}
\label{sec:conclusion}

We present a unified empirical study of group-wise preference optimization for LLMs. 
Across offline and online settings, we show that leveraging multiple responses per prompt consistently outperforms pairwise training, and that positive-sample NLL loss is critical for improving the training stability.
The comparable performance of representative group-wise objectives shows that these gains are robust to the specific loss: the more transferable design principles are instead to use multiple responses and to regularize the positive-response likelihood.

To address the main computational bottleneck of group-wise training, we develop a memory-efficient surrogate that specializes first-order linearization to group-wise preference objectives. It avoids joint backpropagation over response groups while preserving first-order gradient equivalence, substantially reducing peak memory overhead and thereby enabling training with larger group sizes.

Overall, our results suggest that group-wise preference optimization is both effective and scalable when implemented efficiently, enabling richer group-level supervision for future post-training.

\section*{Limitations}
While our proposed algorithm significantly reduces the memory requirements associated with scaling group size (e.g., maintaining constant memory with respect to group size), it necessitates an additional no-grad forward pass. Although this requirement is generally acceptable, it inevitably introduces a trade-off between latency and memory efficiency. 
Furthermore, because our surrogate objective is a first-order linearization of the original DPO loss, it preserves parameter gradients for first-order optimization methods (e.g., SGD or AdamW)~\citep{robbins1951stochastic, loshchilov2017decoupled}, but does not retain the second-order structure. 
As a result, optimization algorithms that rely on higher-order information may have different behaviors.

\bibliography{custom}

\clearpage
\appendix
\section{Experimental Details}\label{appendix:experimental_details}
This section provides additional details on the experimental setup, including the data preparation, benchmarks and training / eval hyperparameters.

\subsection{Offline Data}
For offline training, we sample prompts from \href{https://huggingface.co/datasets/allenai/Dolci-Instruct-DPO}{Dolci-Instruct-DPO}~\citep{olmo2025olmo}.
Before generation, we filter the data to ensure that the tokenized prompt length does not exceed 2048 under the tokenizations of all participating models.
For each remaining prompt, we generate multiple candidate responses using a collection of models, as summarized in Table~\ref{tab:appendix_models}.
These responses are then scored using \href{https://huggingface.co/Skywork/Skywork-Reward-V2-Qwen3-8B}{Skywork-Reward-V2-Qwen3-8B}.
Based on the reward scores, responses are partitioned into two groups: the top-$k$ responses constitute the positive group, while the bottom-$k$ responses form the negative group.
To ensure a clear separation between groups, we discard prompts for which the boundary scores overlap.
After grouping, the scores are discarded, and responses within each group are treated as unordered, with no ranking.
\begin{table}[ht]
\centering
\vspace{-0.2em}
\renewcommand{\arraystretch}{1.05}
\resizebox{\columnwidth}{!}{
\begin{tabular}{c}
\hline
\textbf{Model} \\
\hline
\href{https://huggingface.co/google/gemma-3-1b-it}{google/gemma-3-1b-it}~\citep{gemmateam2025gemma3technicalreport} \\
\href{https://huggingface.co/google/gemma-3-4b-it}{google/gemma-3-4b-it}~\citep{gemmateam2025gemma3technicalreport} \\
\href{https://huggingface.co/google/gemma-3-12b-it}{google/gemma-3-12b-it}~\citep{gemmateam2025gemma3technicalreport} \\
\href{https://huggingface.co/google/gemma-3-27b-it}{google/gemma-3-27b-it}~\citep{gemmateam2025gemma3technicalreport} \\
\href{https://huggingface.co/allenai/Olmo-3-7B-Instruct}{allenai/Olmo-3-7B-Instruct}~\citep{olmo2025olmo} \\
\href{https://huggingface.co/allenai/Olmo-3.1-32B-Instruct}{allenai/Olmo-3.1-32B-Instruct}~\citep{olmo2025olmo} \\
\href{https://huggingface.co/openai/gpt-oss-20b}{openai/gpt-oss-20b}~\citep{agarwal2025gpt} \\
\href{https://huggingface.co/openai/gpt-oss-120b}{openai/gpt-oss-120b}~\citep{agarwal2025gpt} \\
\href{https://huggingface.co/microsoft/Phi-4-mini-instruct}{microsoft/Phi-4-mini-instruct}~\citep{abouelenin2025phi} \\
\href{https://huggingface.co/Qwen/Qwen3-0.6B}{Qwen/Qwen3-0.6B}~\citep{yang2025qwen3} \\
\href{https://huggingface.co/Qwen/Qwen3-1.7B}{Qwen/Qwen3-1.7B}~\citep{yang2025qwen3} \\
\href{https://huggingface.co/Qwen/Qwen3-4B}{Qwen/Qwen3-4B}~\citep{yang2025qwen3} \\
\href{https://huggingface.co/Qwen/Qwen3-14B}{Qwen/Qwen3-14B}~\citep{yang2025qwen3} \\
\href{https://huggingface.co/Qwen/Qwen3-32B}{Qwen/Qwen3-32B}~\citep{yang2025qwen3} \\
\href{https://huggingface.co/Qwen/Qwen3-30B-A3B-Instruct-2507}{Qwen/Qwen3-30B-A3B-Instruct-2507}~\citep{yang2025qwen3} \\
\href{https://huggingface.co/01-ai/Yi-1.5-34B-Chat-16K}{01-ai/Yi-1.5-34B-Chat-16K}~\citep{young2024yi} \\
\hline
\end{tabular}
}
\vspace{-0.5em}
\caption{Models used for response generation.}
\label{tab:appendix_models}
\end{table}

\subsection{Benchmarks and Evaluation}\label{appendix:experimental_details_dataset}
We summarize all evaluation benchmarks and their corresponding generation configurations in Table~\ref{tab:eval_config}.

\subsection{Hyperparameters}
All experiments are conducted on NVIDIA H100 (80GB) SXM GPUs. Table~\ref{tab:hyperparameters} summarizes the hyperparameters used in both offline and online training. In the online setting, we adopt asynchronous training: 8 GPUs for training and 16 GPUs for rollout.
\begin{table}[ht]
\centering
\footnotesize
\setlength{\tabcolsep}{3pt}
\renewcommand{\arraystretch}{1.08}

\begin{tabular}{@{}
>{\raggedright\arraybackslash}m{0.58\columnwidth}
>{\centering\arraybackslash}m{0.15\columnwidth}
>{\centering\arraybackslash}m{0.10\columnwidth}
>{\centering\arraybackslash}m{0.09\columnwidth}
@{}}
\toprule
Benchmarks & Resp. Len. & Temp. & Top-$p$ \\
\midrule

AIME24 / AIME25 / AMC23 / GPQA-D~\citep{rein2024gpqa}
& 8192 & 1.0 & 0.7 \\
\midrule

Math500~\citep{lightman2023let} /
Minerva~\citep{lewkowycz2022solving} /
OlympiadBench~\citep{he2024olympiadbench} /
MMLU-Pro~\citep{wang2024mmlu} /
IfEval~\citep{zhou2023instruction} /
IfBench~\citep{pyatkin2025generalizing} /
ZebraLogic~\citep{lin2025zebralogic} /
BBH~\citep{srivastava2023beyond} /
AGIEval Eng~\citep{zhong2024agieval}
& 8192 & 0.0 & 1.0 \\
\midrule

HumanEval+~\citep{liu2023your} /
LiveCodeBench v6~\citep{jain2024livecodebench}
& 8192 & 0.6 & 0.95 \\
\bottomrule
\end{tabular}
\caption{Generation configs for different benchmarks.}
\label{tab:eval_config}
\end{table}
\begin{table}[ht]
\centering
\footnotesize
\setlength{\tabcolsep}{2pt}
\renewcommand{\arraystretch}{1.1}
\resizebox{\columnwidth}{!}{
\begin{tabular}{lcccc}
\toprule
\multirow{2}{*}{Hyperparameter} & gemma-3-4b & olmo-3-7b & olmo-3.1-32b & qwen3-4b \\
\cmidrule(lr){2-5}
 & offline & offline & offline & online \\
\midrule
Backend      & fsdp2 & fsdp2 & fsdp2 & fsdp2 \\
Rollout      & vllm & vllm & vllm & vllm \\
Learning Rate & 1e-6 & 1e-6 & 1e-6 & 1e-6 \\
Batch Size & 128 & 128 & 128 & 512 \\
Mini Batch Size & 128 & 128 & 128 & 64 \\
Num Responses & – & – & – & 8 \\
LR Scheduler & linear & linear & linear & constant \\
Weight Decay & 0.0 & 0.0 & 0.0 & 0.0 \\
$\beta$ & 2/5/10/15 & 2/5/10/15 & 2 & 5 \\
nll-coeff & 0/1 & 0/1 & 0/1 & 0/1 \\
Log-Probs Agg & mean & mean & mean & mean \\
Prompt Len & 2048 & 2048 & 2048 & 2048 \\
Response Len & 8192 & 8192 & 8192 & 4096 \\
Num Epochs & 1 & 1 & 1 & – \\
Num Steps & 1139 & 1139 & 1139 & 100 \\
\midrule
Num GPUs & 8 & 8 & 32 & 8T + 16R \\
\bottomrule
\end{tabular}
}
\caption{Offline and online DPO hyperparameters. Values separated by “/” denote hyperparameter sweeps.}
\label{tab:hyperparameters}
\end{table}

\section{Additional Analyses}
\subsection{Efficiency Analysis}
\begin{figure*}[htbp]
  \centering
  \begin{subfigure}{\linewidth}
    \centering
    \includegraphics[width=\linewidth]{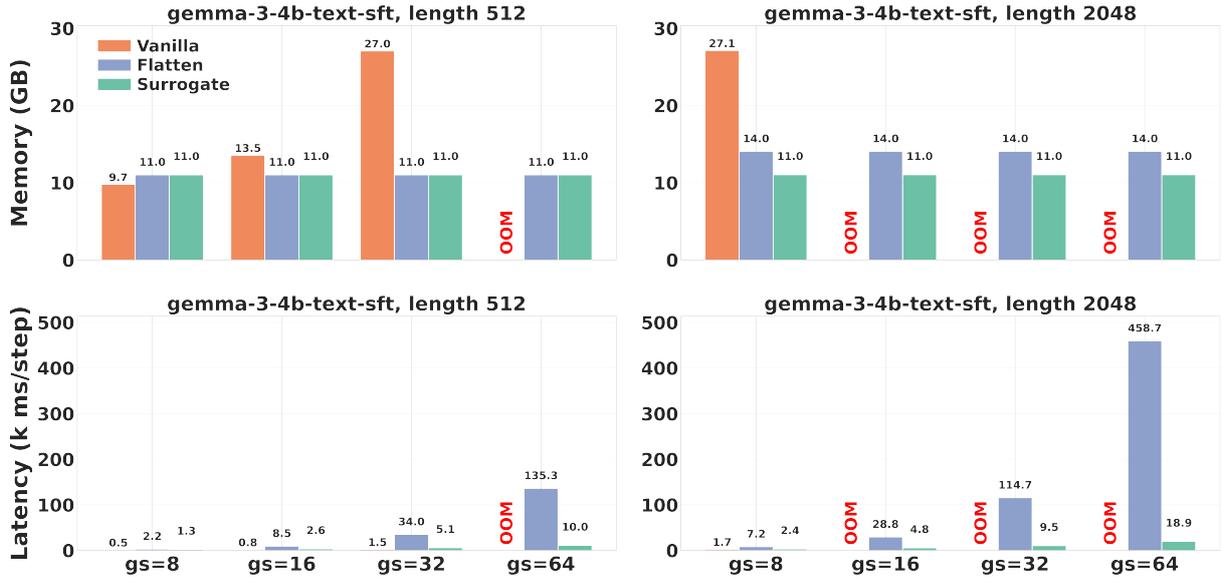}
  \end{subfigure}

    \caption{Peak memory overhead and average step latency for \texttt{gemma-3-4b-sft} across sequence lengths and group sizes. The surrogate significantly reduces memory vs.\ the vanilla implementation with modest latency overhead.}
  \label{fig:appendix_efficiency}
\end{figure*}

In Figure~\ref{fig:efficiency}, we present the memory and latency comparison for \texttt{olmo-3-7b-it-sft}. Here, we additionally include results for \texttt{gemma-3-4b-sft}, as shown in Figure~\ref{fig:appendix_efficiency}.
Overall, the conclusions remain consistent: our surrogate implementation significantly reduces memory usage and maintains constant memory overhead across group sizes, at the cost of a modest increase in latency.

\subsection{Training-Dynamics Diagnostics}
\label{appendix:training_dynamics}
\begin{figure*}[htbp]
  \centering
  \includegraphics[width=\linewidth]{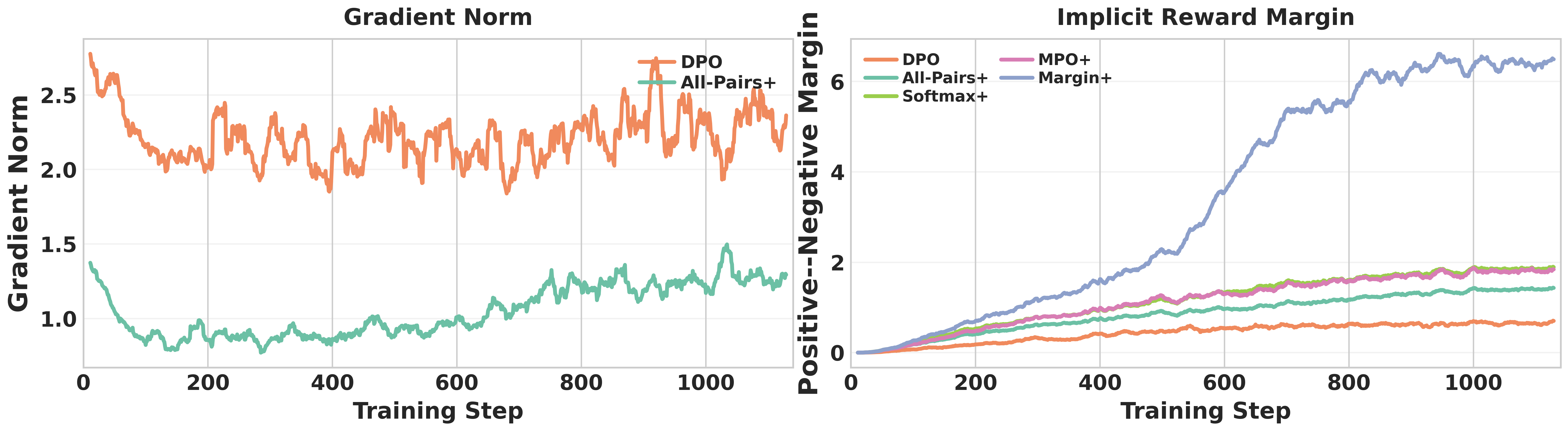}
  \caption{Training-dynamics diagnostics for random-pair DPO and group-wise objectives on \texttt{olmo-3-7b-it-sft}. All methods use $\beta=2$ and positive-response NLL. Left: All-Pairs has a lower and more stable gradient norm than random-pair DPO. Right: the group-wise objectives can learn larger positive-negative implicit reward margins.}
  \label{fig:training_dynamics_diagnostics}
\end{figure*}

\begin{figure*}[ht]
  \centering
  \includegraphics[width=\linewidth]{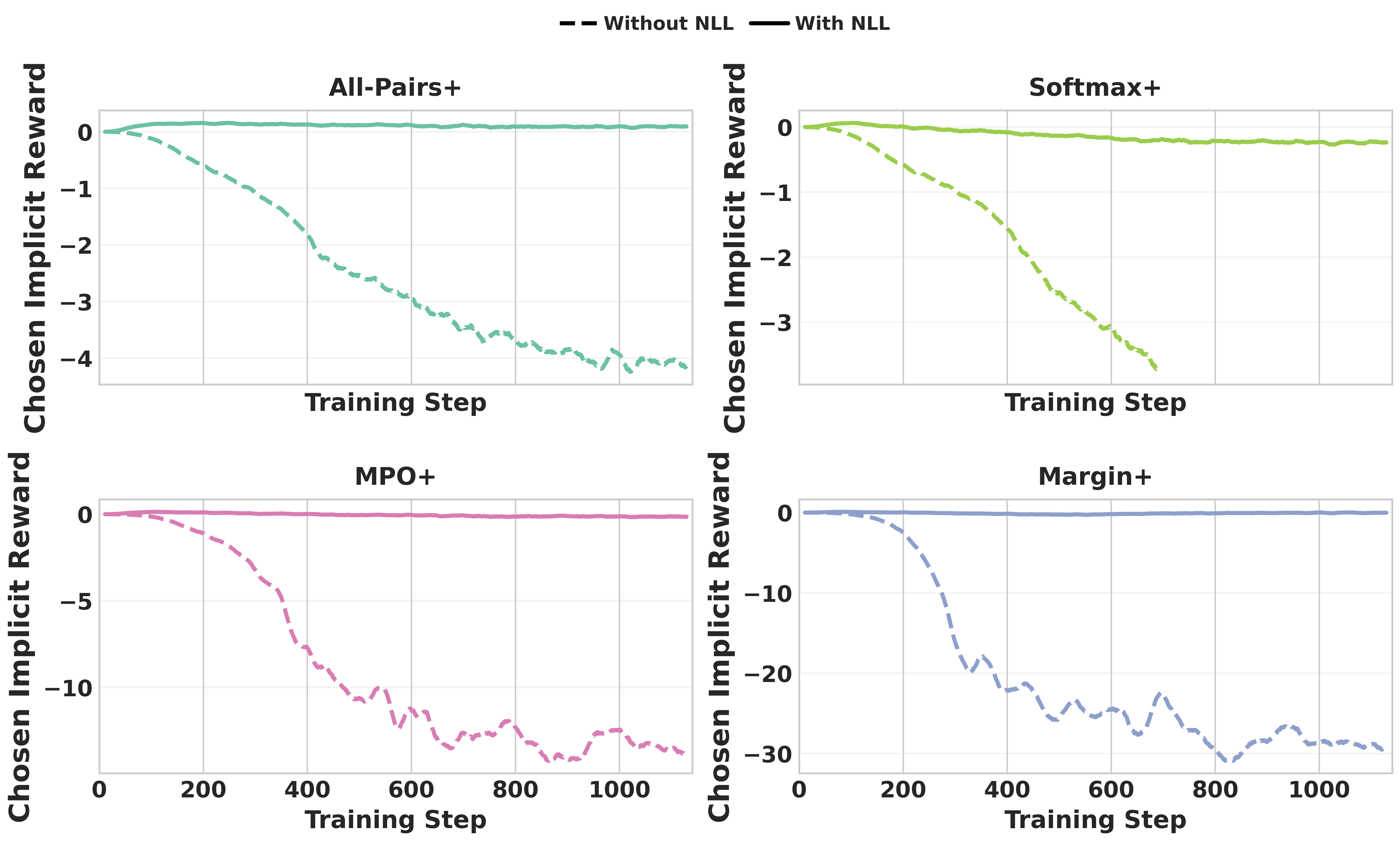}
  \caption{Chosen implicit reward with and without positive-response NLL for four group-wise objectives on \texttt{olmo-3-7b-it-sft} with $\beta=2$. Without NLL, the chosen reward drops sharply; NLL largely prevents this.}
  \label{fig:chosen_reward_nll_diagnostics}
\end{figure*}

Figures~\ref{fig:training_dynamics_diagnostics} and~\ref{fig:chosen_reward_nll_diagnostics} show offline dynamics for \texttt{olmo-3-7b-it-sft} at $1e\!-\!6$ and $\beta=2$.
Curves are 21-step moving averages for readability.

Figure~\ref{fig:training_dynamics_diagnostics} complements the variance-reduction analysis in Section~\ref{sec:groupisbetter}.
All-Pairs has a lower and more stable gradient norm than random-pair DPO, which is consistent with averaging over the arbitrary choice of a single positive-negative pair.
The gradient norm is not itself a direct estimate of gradient variance, but its reduced magnitude and fewer large spikes agree with the theoretical interpretation.
Moreover, all four group-wise objectives learn larger positive-negative implicit reward margins than random-pair DPO under the same value of $\beta$.

Figure~\ref{fig:chosen_reward_nll_diagnostics} further explains the stabilizing effect of positive-response NLL.
Without NLL, the chosen implicit reward decreases substantially during training for every evaluated group-wise objective, indicating degradation in positive-response likelihood relative to the reference policy.
Adding NLL largely prevents this decrease, providing a training-dynamics explanation for the improved stability and final performance observed in Section~\ref{sec:nll_stability}.

\section{Surrogate--Vanilla Sanity Check}
\label{appendix:surrogate_sanity}
The surrogate should be viewed as a gradient carrier rather than a scalar approximation whose value must match the group objective away from the current parameter point.
Because its coefficients are recomputed at every optimization step, the surrogate and vanilla implementations have the same first-order gradient when evaluated at the same parameters.
Thus, the linearization itself does not introduce accumulating systematic approximation error for first-order optimization; any discrepancy only arises from implementation-level effects such as finite-precision arithmetic and batching order.

To quantify the numerical difference between our proposed surrogate and the vanilla implementation empirically, we ran an additional full-parameter Qwen3-0.6B-Base sanity check in a small setting where vanilla full-group DPO is feasible. We compare the surrogate and vanilla implementations over 10 optimizer steps on synthetic random-token batches with 5 groups. All experiments are run in bf16, matching our main training setup, so the reported discrepancies include numerical noise. We report both step-1 gradient diagnostics and multi-step parameter-update diagnostics. Specifically, ``Min step-1 grad cos'' is the minimum cosine similarity between the surrogate and vanilla gradients at the first step over group sizes 2/4/8, and ``Max step-1 grad RMS'' is the maximum RMS gradient difference over these group sizes. These step-1 metrics measure agreement when both implementations are evaluated at the same parameters. We then report the RMS difference between the final parameter updates after 10 steps for each group size, together with the maximum absolute coordinate-wise update difference. We test four objectives (All-Pairs, MPO, Margin, Softmax), group sizes 2/4/8, two optimizers (AdamW and Muon), and commonly used learning rates $1e-5$ and $1e-6$.

\begin{table*}[htbp]
\centering
\scriptsize
\setlength{\tabcolsep}{3.5pt}
\resizebox{\linewidth}{!}{%
\begin{tabular}{lllrrrrrr}
\toprule
Objective & Optimizer & LR & Grad cos $\uparrow$ & Grad RMS $\downarrow$ & \multicolumn{3}{c}{Update RMS $\downarrow$} & Max update abs $\downarrow$ \\
\cmidrule(lr){6-8}
 & & & & & gs=2 & gs=4 & gs=8 & \\
\midrule
All-Pairs & AdamW & $1e\!-\!5$ & 0.995013 & 2.67e-5 & 2.71e-6 & 3.21e-6 & 3.55e-6 & 2.59e-4 \\
All-Pairs & AdamW & $1e\!-\!6$ & 0.993933 & 6.02e-5 & 2.47e-7 & 2.33e-7 & 2.26e-7 & 2.48e-5 \\
All-Pairs & Muon  & $1e\!-\!5$ & 0.995756 & 3.46e-5 & 7.36e-8 & 6.59e-8 & 5.75e-8 & 2.44e-4 \\
All-Pairs & Muon  & $1e\!-\!6$ & 0.959863 & 1.23e-4 & 2.26e-9 & 2.51e-9 & 2.25e-9 & 1.72e-5 \\
\midrule
MPO & AdamW & $1e\!-\!5$ & 0.989735 & 5.02e-5 & 2.87e-6 & 3.57e-6 & 4.09e-6 & 2.44e-4 \\
MPO & AdamW & $1e\!-\!6$ & 0.997864 & 2.65e-5 & 2.18e-7 & 2.25e-7 & 2.27e-7 & 2.67e-5 \\
MPO & Muon  & $1e\!-\!5$ & 0.994669 & 3.81e-5 & 7.71e-8 & 6.68e-8 & 6.39e-8 & 1.83e-4 \\
MPO & Muon  & $1e\!-\!6$ & 0.992146 & 5.50e-5 & 2.77e-9 & 2.26e-9 & 2.34e-9 & 1.72e-5 \\
\midrule
Margin & AdamW & $1e\!-\!5$ & 0.995498 & 3.25e-5 & 3.26e-6 & 2.89e-6 & 3.50e-6 & 2.14e-4 \\
Margin & AdamW & $1e\!-\!6$ & 0.993279 & 5.15e-5 & 2.44e-7 & 2.27e-7 & 2.29e-7 & 2.96e-5 \\
Margin & Muon  & $1e\!-\!5$ & 0.995170 & 4.25e-5 & 6.53e-8 & 6.67e-8 & 6.29e-8 & 1.37e-4 \\
Margin & Muon  & $1e\!-\!6$ & 0.982327 & 7.22e-5 & 2.65e-9 & 2.48e-9 & 2.15e-9 & 1.91e-5 \\
\midrule
Softmax & AdamW & $1e\!-\!5$ & 0.987822 & 5.20e-5 & 3.26e-6 & 3.50e-6 & 3.54e-6 & 2.44e-4 \\
Softmax & AdamW & $1e\!-\!6$ & 0.995575 & 4.36e-5 & 2.22e-7 & 2.26e-7 & 2.23e-7 & 2.67e-5 \\
Softmax & Muon  & $1e\!-\!5$ & 0.994778 & 4.03e-5 & 7.17e-8 & 6.67e-8 & 6.03e-8 & 1.68e-4 \\
Softmax & Muon  & $1e\!-\!6$ & 0.989825 & 5.25e-5 & 2.30e-9 & 2.23e-9 & 1.98e-9 & 1.14e-5 \\
\bottomrule
\end{tabular}%
}
\caption{Numerical agreement between the surrogate and vanilla implementations across group sizes. ``Grad cos'' is the minimum step-1 gradient cosine similarity over group sizes of 2/4/8, and ``Grad RMS'' is the maximum step-1 RMS gradient difference. Update RMS is measured after 10 optimizer steps to capture accumulated parameter drift.}
\label{tab:surrogate_sanity}
\end{table*}

The two implementations agree closely across all tested settings.
At step 1, the minimum gradient cosine similarity is 0.9599, with most values above 0.98, while the maximum RMS gradient difference is $1.23e\!-\!4$.
After 10 steps, update RMS differences remain on the order of $1e\!-\!6$ for AdamW and $1e\!-\!8$ for Muon at learning rate $1e\!-\!5$, and decrease further at $1e\!-\!6$.
We do not observe a monotonic increase in discrepancy as the group size increases.
The results show that surrogate introduces negligible parameter drift across the tested first-order settings.

\section{Full Experimental Results}
\label{appendix:full_results}
We present results for all evaluated checkpoints, including $\beta$ and NLL coefficient sweeps, in Table~\ref{tab:dpo-edge} and the subsequent figures through Figure~\ref{fig:full_results_end}.
\begin{table*}[ht]
\centering
\scriptsize
\setlength{\tabcolsep}{1.8pt}
\renewcommand{\arraystretch}{1.20}
\resizebox{\linewidth}{!}{%
\begin{tabularx}{\linewidth}{@{}>{\raggedright\arraybackslash}Xcccccc@{}}
\toprule
\multicolumn{7}{c}{\textbf{\textit{DPO$^{\dagger}$ and RFT}}} \\
\multirow{4}{*}{Dataset} & \multicolumn{2}{c}{gemma-3-4b-sft} & \multicolumn{2}{c}{olmo-3-7b-it-sft} & \multicolumn{2}{c}{olmo-3.1-32b-it-sft} \\
\cmidrule(lr){2-3} \cmidrule(lr){4-5} \cmidrule(lr){6-7}
 & \multicolumn{1}{c}{DPO$^{\dagger}$} & \multicolumn{1}{c}{RFT} & \multicolumn{1}{c}{DPO$^{\dagger}$} & \multicolumn{1}{c}{RFT} & \multicolumn{1}{c}{DPO$^{\dagger}$} & \multicolumn{1}{c}{RFT} \\
\cmidrule(lr){2-7}
 & nll=1.0, $\beta$=2.0 & - & nll=1.0, $\beta$=2.0 & - & nll=1.0, $\beta$=2.0 & - \\
\cmidrule(lr){2-7}
 & 200 / 1139 & 1139 / 1139 & 500 / 1139 & 1000 / 1139 & 900 / 1139 & 700 / 1139 \\
\midrule
\addlinespace[2pt]
\rowcolor{tablepink}\multicolumn{7}{@{}l}{\textbf{Math}}\\
\addlinespace[2pt]
\quad AIME24 & - & - & 32.7 / 30.8 & 15.8 / 15.8 & 45.4 / 42.7 & 27.1 / 29.4 \\
\quad AIME25 & - & - & 24.4 / 22.3 & 17.3 / 15.8 & 33.5 / 35.8 & 27.5 / 23.7 \\
\quad AMC23 & 16.7 / 16.4 & 14.7 / 14.7 & 70.3 / 70.6 & 54.7 / 53.8 & 78.8 / 77.3 & 68.0 / 66.4 \\
\quad MATH500 & 38.2 / 37.6 & 38.4 / 38.4 & 85.8 / 84.0 & 77.6 / 75.4 & 90.2 / 90.8 & 83.0 / 83.2 \\
\quad Minerva & 18.0 / 16.2 & 16.9 / 16.9 & 36.0 / 34.6 & 33.1 / 32.0 & 44.5 / 42.6 & 43.0 / 39.0 \\
\quad Olympiad & 12.6 / 12.6 & 11.0 / 11.0 & 57.5 / 54.5 & 43.7 / 45.5 & 64.1 / 62.7 & 52.9 / 52.9 \\
\textbf{Math Avg} & 21.4 / 20.7 & 20.2 / 20.2 & 51.1 / 49.5 & 40.4 / 39.7 & \textbf{\textit{59.4 / 58.7}} & 50.2 / 49.1 \\
\addlinespace[2pt]
\rowcolor{tablepink}\multicolumn{7}{@{}l}{\textbf{Reasoning}}\\
\addlinespace[2pt]
\quad AGIEval Eng & 51.1 / 50.1 & 49.0 / 49.0 & 60.1 / 61.1 & 55.8 / 55.2 & 73.7 / 74.1 & 69.0 / 69.7 \\
\quad GPQA-D & 21.8 / 23.4 & 21.4 / 21.4 & 35.5 / 36.9 & 27.9 / 28.3 & 46.7 / 46.8 & 33.2 / 34.5 \\
\quad MMLU-PRO & 35.0 / 34.4 & 32.8 / 32.8 & 54.7 / 54.3 & 46.3 / 46.6 & 67.5 / 67.5 & 58.2 / 58.6 \\
\quad ZebraLogic & 19.1 / 16.7 & 11.7 / 11.7 & 21.4 / 28.1 & 15.8 / 15.9 & 19.4 / 18.3 & 28.3 / 28.1 \\
\quad BBH & 50.0 / 49.8 & 49.9 / 49.9 & 73.6 / 73.8 & 66.7 / 66.6 & 80.8 / 80.9 & 77.6 / 77.7 \\
\addlinespace[2pt]
\rowcolor{tablepink}\multicolumn{7}{@{}l}{\textbf{Coding}}\\
\addlinespace[2pt]
\quad HumanEval+ & 55.2 / 47.9 & 56.8 / 56.8 & 71.0 / 70.9 & 72.0 / 72.0 & 83.8 / 84.3 & 83.5 / 81.1 \\
\quad LiveCode v6 & 10.6 / 11.4 & 10.7 / 10.7 & 20.7 / 18.3 & 18.7 / 19.4 & 26.6 / 25.7 & 25.3 / 25.0 \\
\addlinespace[2pt]
\rowcolor{tablepink}\multicolumn{7}{@{}l}{\textbf{Instruction Following}}\\
\addlinespace[2pt]
\quad IFBench & 21.1 / 21.1 & 16.0 / 16.0 & 30.6 / 31.0 & 21.1 / 21.8 & 34.7 / 35.4 & 24.1 / 26.2 \\
\quad IFEval & 62.7 / 56.7 & 58.4 / 58.4 & 83.7 / 81.5 & 73.4 / 73.2 & 86.5 / 86.0 & 83.0 / 82.3 \\
\textbf{General Avg} & 36.3 / 34.6 & 34.1 / 34.1 & 50.2 / 50.6 & 44.2 / 44.3 & \textbf{\textit{57.7 / 57.7}} & 53.6 / 53.7 \\
\midrule
\textbf{Avg} & 31.7 / 30.3 & 29.8 / 29.8 & 50.5 / 50.2 & 42.7 / 42.5 & \textbf{\textit{58.4 / 58.1}} & 52.2 / 51.8 \\
\bottomrule
\end{tabularx}%
}
\caption{Performance of DPO$^{\dagger}$ and RFT across models, varying $\beta$ and nll. DPO$^{\dagger}$ constructs pairs by using samples with the highest reward as positives and samples with the lowest reward as negatives.}
\label{tab:dpo-edge}
\end{table*}

\begin{table*}[ht]
\centering
\scriptsize
\setlength{\tabcolsep}{1.8pt}
\renewcommand{\arraystretch}{1.20}
\resizebox{\linewidth}{!}{%
\begin{tabularx}{\linewidth}{@{}>{\raggedright\arraybackslash}Xccccccc@{}}
\toprule
\multicolumn{8}{c}{\textbf{\textit{gemma-3-4b-sft}}} \\
\multirow{4}{*}{Dataset} & \multicolumn{7}{c}{DPO} \\
\cmidrule(lr){2-8}
 & \multicolumn{1}{c}{nll=0.0} & \multicolumn{6}{c}{nll=1.0} \\
\cmidrule(lr){2-2} \cmidrule(lr){3-8}
 & \multicolumn{1}{c}{$\beta$=2.0} & \multicolumn{2}{c}{$\beta$=2.0} & \multicolumn{1}{c}{$\beta$=5.0} & \multicolumn{1}{c}{$\beta$=10.0} & \multicolumn{2}{c}{$\beta$=15.0} \\
\cmidrule(lr){2-2} \cmidrule(lr){3-4} \cmidrule(lr){5-5} \cmidrule(lr){6-6} \cmidrule(lr){7-8}
 & 1139 & 800 & 1139 & 1139 & 1139 & 200 & 1139 \\
\midrule
\addlinespace[2pt]
\rowcolor{tablepink}\multicolumn{8}{@{}l}{\textbf{Math}}\\
\addlinespace[2pt]
\quad AIME24 & - & - & - & - & - & - & - \\
\quad AIME25 & - & - & - & - & - & - & - \\
\quad AMC23 & 6.2 & 17.0 & 15.6 & 14.4 & 16.2 & 15.3 & 16.4 \\
\quad MATH500 & 26.0 & 40.4 & 40.8 & 39.8 & 40.2 & 39.0 & 36.2 \\
\quad Minerva & 5.9 & 19.1 & 17.3 & 21.0 & 18.4 & 20.6 & 15.4 \\
\quad Olympiad & 5.2 & 12.6 & 12.1 & 12.0 & 10.8 & 11.6 & 12.9 \\
\textbf{Math Avg} & 10.8 & \textbf{\textit{22.3}} & 21.5 & 21.8 & 21.4 & 21.6 & 20.2 \\
\addlinespace[2pt]
\rowcolor{tablepink}\multicolumn{8}{@{}l}{\textbf{Reasoning}}\\
\addlinespace[2pt]
\quad AGIEval Eng & 48.7 & 50.3 & 51.4 & 52.3 & 50.3 & 50.6 & 51.0 \\
\quad GPQA-D & 25.8 & 21.3 & 20.6 & 22.1 & 26.3 & 26.7 & 26.9 \\
\quad MMLU-PRO & 31.9 & 34.1 & 33.5 & 34.5 & 36.7 & 35.7 & 36.0 \\
\quad ZebraLogic & 7.9 & 15.8 & 15.9 & 17.6 & 18.9 & 17.3 & 17.5 \\
\quad BBH & 40.5 & 50.1 & 51.3 & 50.6 & 50.5 & 45.6 & 47.0 \\
\addlinespace[2pt]
\rowcolor{tablepink}\multicolumn{8}{@{}l}{\textbf{Coding}}\\
\addlinespace[2pt]
\quad HumanEval+ & 28.2 & 57.7 & 56.5 & 58.4 & 57.5 & 54.5 & 56.8 \\
\quad LiveCode v6 & 6.1 & 11.4 & 11.9 & 12.1 & 10.6 & 10.3 & 9.9 \\
\addlinespace[2pt]
\rowcolor{tablepink}\multicolumn{8}{@{}l}{\textbf{Instruction Following}}\\
\addlinespace[2pt]
\quad IFBench & 24.5 & 19.7 & 20.7 & 21.8 & 22.8 & 19.0 & 23.1 \\
\quad IFEval & 56.4 & 64.1 & 59.5 & 65.2 & 66.7 & 64.0 & 67.7 \\
\textbf{General Avg} & 30.0 & 36.1 & 35.7 & 37.2 & \textbf{\textit{37.8}} & 36.0 & 37.3 \\
\midrule
\textbf{Avg} & 24.1 & 31.8 & 31.3 & 32.4 & \textbf{\textit{32.8}} & 31.6 & 32.1 \\
\bottomrule
\end{tabularx}%
}
\caption{Performance of DPO on gemma-3-4b-sft with group size 1, varying $\beta$ and nll.}
\end{table*}

\begin{table*}[ht]
\centering
\scriptsize
\setlength{\tabcolsep}{1.8pt}
\renewcommand{\arraystretch}{1.20}
\resizebox{\linewidth}{!}{%
\begin{tabularx}{\linewidth}{@{}>{\raggedright\arraybackslash}Xccccccccc@{}}
\toprule
\multicolumn{10}{c}{\textbf{\textit{gemma-3-4b-sft}}} \\
\multirow{4}{*}{Dataset} & \multicolumn{9}{c}{Margin DPO+} \\
\cmidrule(lr){2-10}
 & \multicolumn{1}{c}{nll=0.0} & \multicolumn{8}{c}{nll=1.0} \\
\cmidrule(lr){2-2} \cmidrule(lr){3-10}
 & \multicolumn{1}{c}{$\beta$=2.0} & \multicolumn{2}{c}{$\beta$=2.0} & \multicolumn{2}{c}{$\beta$=5.0} & \multicolumn{2}{c}{$\beta$=10.0} & \multicolumn{2}{c}{$\beta$=15.0} \\
\cmidrule(lr){2-2} \cmidrule(lr){3-4} \cmidrule(lr){5-6} \cmidrule(lr){7-8} \cmidrule(lr){9-10}
 & 1139 & 100 & 1139 & 900 & 1139 & 900 & 1139 & 600 & 1139 \\
\midrule
\addlinespace[2pt]
\rowcolor{tablepink}\multicolumn{10}{@{}l}{\textbf{Math}}\\
\addlinespace[2pt]
\quad AIME24 & - & - & - & - & - & - & - & - & - \\
\quad AIME25 & - & - & - & - & - & - & - & - & - \\
\quad AMC23 & 0.8 & 16.7 & 14.2 & 15.0 & 15.2 & 16.1 & 14.4 & 16.2 & 14.4 \\
\quad MATH500 & 2.8 & 40.4 & 35.8 & 40.6 & 36.0 & 40.4 & 38.0 & 41.6 & 39.4 \\
\quad Minerva & 0.7 & 17.3 & 18.0 & 18.0 & 18.4 & 20.2 & 17.6 & 18.0 & 21.7 \\
\quad Olympiad & 0.4 & 11.9 & 11.0 & 12.9 & 11.9 & 13.3 & 13.9 & 12.4 & 11.3 \\
\textbf{Math Avg} & 1.2 & 21.6 & 19.7 & 21.6 & 20.3 & \textbf{\textit{22.5}} & 21.0 & 22.1 & 21.7 \\
\addlinespace[2pt]
\rowcolor{tablepink}\multicolumn{10}{@{}l}{\textbf{Reasoning}}\\
\addlinespace[2pt]
\quad AGIEval Eng & 12.3 & 50.5 & 48.1 & 48.0 & 48.1 & 50.3 & 50.1 & 50.1 & 50.1 \\
\quad GPQA-D & 0.4 & 24.1 & 22.6 & 21.7 & 22.9 & 24.0 & 23.1 & 24.0 & 23.5 \\
\quad MMLU-PRO & 0.4 & 34.7 & 33.4 & 33.9 & 34.3 & 34.9 & 34.6 & 35.2 & 34.7 \\
\quad ZebraLogic & 0.0 & 16.9 & 13.7 & 15.3 & 16.0 & 15.4 & 15.2 & 16.2 & 15.7 \\
\quad BBH & 24.3 & 49.3 & 47.6 & 48.7 & 48.5 & 49.7 & 49.5 & 49.5 & 49.9 \\
\addlinespace[2pt]
\rowcolor{tablepink}\multicolumn{10}{@{}l}{\textbf{Coding}}\\
\addlinespace[2pt]
\quad HumanEval+ & 2.7 & 58.0 & 50.1 & 53.4 & 52.1 & 55.7 & 55.4 & 56.2 & 54.6 \\
\quad LiveCode v6 & 0.0 & 11.5 & 10.8 & 11.8 & 11.8 & 11.5 & 11.0 & 11.6 & 12.2 \\
\addlinespace[2pt]
\rowcolor{tablepink}\multicolumn{10}{@{}l}{\textbf{Instruction Following}}\\
\addlinespace[2pt]
\quad IFBench & 24.8 & 21.1 & 20.1 & 20.1 & 17.7 & 19.0 & 20.1 & 20.1 & 19.4 \\
\quad IFEval & 41.6 & 63.6 & 61.2 & 60.8 & 62.7 & 60.3 & 59.0 & 60.6 & 59.9 \\
\textbf{General Avg} & 11.8 & \textbf{\textit{36.6}} & 34.2 & 34.8 & 34.9 & 35.7 & 35.3 & 36.0 & 35.5 \\
\midrule
\textbf{Avg} & 8.6 & \textbf{\textit{32.0}} & 29.7 & 30.8 & 30.4 & 31.6 & 30.9 & 31.7 & 31.3 \\
\bottomrule
\end{tabularx}%
}
\caption{Performance of Margin DPO+ on gemma-3-4b-sft with group size 8, varying $\beta$ and nll.}
\end{table*}

\begin{table*}[ht]
\centering
\scriptsize
\setlength{\tabcolsep}{1.8pt}
\renewcommand{\arraystretch}{1.20}
\resizebox{\linewidth}{!}{%
\begin{tabularx}{\linewidth}{@{}>{\raggedright\arraybackslash}Xcccccccc@{}}
\toprule
\multicolumn{9}{c}{\textbf{\textit{gemma-3-4b-sft}}} \\
\multirow{4}{*}{Dataset} & \multicolumn{8}{c}{MPO+} \\
\cmidrule(lr){2-9}
 & \multicolumn{1}{c}{nll=0.0} & \multicolumn{7}{c}{nll=1.0} \\
\cmidrule(lr){2-2} \cmidrule(lr){3-9}
 & \multicolumn{1}{c}{$\beta$=2.0} & \multicolumn{2}{c}{$\beta$=2.0} & \multicolumn{2}{c}{$\beta$=5.0} & \multicolumn{1}{c}{$\beta$=10.0} & \multicolumn{2}{c}{$\beta$=15.0} \\
\cmidrule(lr){2-2} \cmidrule(lr){3-4} \cmidrule(lr){5-6} \cmidrule(lr){7-7} \cmidrule(lr){8-9}
 & 1139 & 900 & 1139 & 600 & 1139 & 1139 & 1100 & 1139 \\
\midrule
\addlinespace[2pt]
\rowcolor{tablepink}\multicolumn{9}{@{}l}{\textbf{Math}}\\
\addlinespace[2pt]
\quad AIME24 & - & - & - & - & - & - & - & - \\
\quad AIME25 & - & - & - & - & - & - & - & - \\
\quad AMC23 & 1.6 & 18.3 & 17.0 & 15.8 & 15.9 & 14.8 & 16.6 & 16.7 \\
\quad MATH500 & 8.0 & 38.6 & 40.4 & 42.8 & 39.6 & 42.6 & 39.6 & 36.8 \\
\quad Minerva & 2.2 & 17.3 & 16.2 & 17.3 & 16.9 & 18.8 & 18.4 & 18.0 \\
\quad Olympiad & 1.5 & 11.9 & 12.0 & 13.8 & 12.9 & 13.3 & 13.2 & 12.9 \\
\textbf{Math Avg} & 3.3 & 21.5 & 21.4 & \textbf{\textit{22.4}} & 21.3 & 22.4 & 21.9 & 21.1 \\
\addlinespace[2pt]
\rowcolor{tablepink}\multicolumn{9}{@{}l}{\textbf{Reasoning}}\\
\addlinespace[2pt]
\quad AGIEval Eng & 46.1 & 49.2 & 49.2 & 50.5 & 49.8 & 50.6 & 51.2 & 50.9 \\
\quad GPQA-D & 19.6 & 20.9 & 19.8 & 22.4 & 22.1 & 22.6 & 25.5 & 25.9 \\
\quad MMLU-PRO & 22.4 & 32.4 & 31.0 & 34.5 & 32.7 & 35.3 & 36.2 & 36.3 \\
\quad ZebraLogic & 0.0 & 15.9 & 16.2 & 16.2 & 16.0 & 15.9 & 16.8 & 15.7 \\
\quad BBH & 38.4 & 50.1 & 49.9 & 50.6 & 50.4 & 52.4 & 51.7 & 52.4 \\
\addlinespace[2pt]
\rowcolor{tablepink}\multicolumn{9}{@{}l}{\textbf{Coding}}\\
\addlinespace[2pt]
\quad HumanEval+ & 0.5 & 53.2 & 53.4 & 49.3 & 49.9 & 50.4 & 52.3 & 52.4 \\
\quad LiveCode v6 & 0.1 & 11.3 & 12.3 & 11.0 & 11.8 & 10.9 & 10.7 & 10.6 \\
\addlinespace[2pt]
\rowcolor{tablepink}\multicolumn{9}{@{}l}{\textbf{Instruction Following}}\\
\addlinespace[2pt]
\quad IFBench & 25.5 & 19.7 & 18.4 & 20.1 & 20.1 & 20.4 & 21.1 & 21.1 \\
\quad IFEval & 37.3 & 55.6 & 57.9 & 57.3 & 54.7 & 64.1 & 66.9 & 65.6 \\
\textbf{General Avg} & 21.1 & 34.3 & 34.2 & 34.7 & 34.2 & 35.8 & \textbf{\textit{36.9}} & 36.8 \\
\midrule
\textbf{Avg} & 15.6 & 30.3 & 30.3 & 30.9 & 30.2 & 31.7 & \textbf{\textit{32.3}} & 31.9 \\
\bottomrule
\end{tabularx}%
}
\caption{Performance of MPO+ on gemma-3-4b-sft with group size 8, varying $\beta$ and nll.}
\end{table*}

\begin{table*}[ht]
\centering
\scriptsize
\setlength{\tabcolsep}{1.8pt}
\renewcommand{\arraystretch}{1.20}
\resizebox{\linewidth}{!}{%
\begin{tabularx}{\linewidth}{@{}>{\raggedright\arraybackslash}Xcccccccc@{}}
\toprule
\multicolumn{9}{c}{\textbf{\textit{gemma-3-4b-sft}}} \\
\multirow{4}{*}{Dataset} & \multicolumn{8}{c}{All-Pairs DPO+} \\
\cmidrule(lr){2-9}
 & \multicolumn{1}{c}{nll=0.0} & \multicolumn{7}{c}{nll=1.0} \\
\cmidrule(lr){2-2} \cmidrule(lr){3-9}
 & \multicolumn{1}{c}{$\beta$=2.0} & \multicolumn{2}{c}{$\beta$=2.0} & \multicolumn{2}{c}{$\beta$=5.0} & \multicolumn{2}{c}{$\beta$=10.0} & \multicolumn{1}{c}{$\beta$=15.0} \\
\cmidrule(lr){2-2} \cmidrule(lr){3-4} \cmidrule(lr){5-6} \cmidrule(lr){7-8} \cmidrule(lr){9-9}
 & 1139 & 600 & 1139 & 900 & 1139 & 1100 & 1139 & 1139 \\
\midrule
\addlinespace[2pt]
\rowcolor{tablepink}\multicolumn{9}{@{}l}{\textbf{Math}}\\
\addlinespace[2pt]
\quad AIME24 & - & - & - & - & - & - & - & - \\
\quad AIME25 & - & - & - & - & - & - & - & - \\
\quad AMC23 & 3.8 & 17.8 & 14.7 & 15.6 & 15.6 & 15.6 & 15.2 & 14.1 \\
\quad MATH500 & 16.8 & 41.2 & 40.2 & 41.4 & 41.8 & 40.0 & 39.4 & 40.0 \\
\quad Minerva & 6.6 & 18.0 & 19.5 & 19.5 & 16.5 & 20.2 & 17.3 & 18.4 \\
\quad Olympiad & 3.4 & 12.7 & 11.4 & 15.4 & 12.6 & 13.5 & 11.9 & 13.3 \\
\textbf{Math Avg} & 7.6 & 22.4 & 21.4 & \textbf{\textit{23.0}} & 21.6 & 22.3 & 20.9 & 21.4 \\
\addlinespace[2pt]
\rowcolor{tablepink}\multicolumn{9}{@{}l}{\textbf{Reasoning}}\\
\addlinespace[2pt]
\quad AGIEval Eng & 42.1 & 50.3 & 49.0 & 49.7 & 50.4 & 51.2 & 50.8 & 51.7 \\
\quad GPQA-D & 23.3 & 23.3 & 21.1 & 22.1 & 21.2 & 23.6 & 23.5 & 25.8 \\
\quad MMLU-PRO & 25.7 & 34.7 & 33.0 & 32.9 & 32.6 & 33.5 & 33.7 & 35.9 \\
\quad ZebraLogic & 0.1 & 16.5 & 16.6 & 14.3 & 14.4 & 16.9 & 15.1 & 16.7 \\
\quad BBH & 39.6 & 50.0 & 50.4 & 50.9 & 51.1 & 50.0 & 50.5 & 46.0 \\
\addlinespace[2pt]
\rowcolor{tablepink}\multicolumn{9}{@{}l}{\textbf{Coding}}\\
\addlinespace[2pt]
\quad HumanEval+ & 1.6 & 53.0 & 52.0 & 46.6 & 46.0 & 48.5 & 49.5 & 51.6 \\
\quad LiveCode v6 & 0.2 & 11.8 & 11.0 & 10.4 & 10.9 & 10.2 & 11.1 & 10.2 \\
\addlinespace[2pt]
\rowcolor{tablepink}\multicolumn{9}{@{}l}{\textbf{Instruction Following}}\\
\addlinespace[2pt]
\quad IFBench & 28.6 & 17.7 & 19.7 & 19.7 & 20.1 & 24.5 & 22.1 & 19.4 \\
\quad IFEval & 36.4 & 54.9 & 54.5 & 55.8 & 55.6 & 65.6 & 65.4 & 66.2 \\
\textbf{General Avg} & 22.0 & 34.7 & 34.2 & 33.6 & 33.6 & \textbf{\textit{36.0}} & 35.7 & 35.9 \\
\midrule
\textbf{Avg} & 17.5 & 30.9 & 30.2 & 30.3 & 29.9 & \textbf{\textit{31.8}} & 31.2 & 31.5 \\
\bottomrule
\end{tabularx}%
}
\caption{Performance of All-Pairs DPO+ on gemma-3-4b-sft with group size 8, varying $\beta$ and nll.}
\end{table*}

\begin{table*}[ht]
\centering
\scriptsize
\setlength{\tabcolsep}{1.8pt}
\renewcommand{\arraystretch}{1.20}
\resizebox{\linewidth}{!}{%
\begin{tabularx}{\linewidth}{@{}>{\raggedright\arraybackslash}Xccccccccc@{}}
\toprule
\multicolumn{10}{c}{\textbf{\textit{gemma-3-4b-sft}}} \\
\multirow{4}{*}{Dataset} & \multicolumn{9}{c}{Softmax DPO+} \\
\cmidrule(lr){2-10}
 & \multicolumn{1}{c}{nll=0.0} & \multicolumn{8}{c}{nll=1.0} \\
\cmidrule(lr){2-2} \cmidrule(lr){3-10}
 & \multicolumn{1}{c}{$\beta$=2.0} & \multicolumn{2}{c}{$\beta$=2.0} & \multicolumn{2}{c}{$\beta$=5.0} & \multicolumn{2}{c}{$\beta$=10.0} & \multicolumn{2}{c}{$\beta$=15.0} \\
\cmidrule(lr){2-2} \cmidrule(lr){3-4} \cmidrule(lr){5-6} \cmidrule(lr){7-8} \cmidrule(lr){9-10}
 & 1139 & 800 & 1139 & 200 & 1139 & 400 & 1139 & 200 & 1139 \\
\midrule
\addlinespace[2pt]
\rowcolor{tablepink}\multicolumn{10}{@{}l}{\textbf{Math}}\\
\addlinespace[2pt]
\quad AIME24 & - & - & - & - & - & - & - & - & - \\
\quad AIME25 & - & - & - & - & - & - & - & - & - \\
\quad AMC23 & 3.4 & 15.9 & 15.3 & 14.5 & 13.3 & 13.8 & 13.6 & 15.5 & 11.6 \\
\quad MATH500 & 15.4 & 38.6 & 35.8 & 40.2 & 36.6 & 36.4 & 32.8 & 36.8 & 35.2 \\
\quad Minerva & 4.8 & 21.0 & 15.1 & 21.3 & 16.5 & 20.2 & 17.6 & 18.0 & 18.0 \\
\quad Olympiad & 2.8 & 12.7 & 13.6 & 10.8 & 12.1 & 9.8 & 11.7 & 11.1 & 11.6 \\
\textbf{Math Avg} & 6.6 & \textbf{\textit{22.1}} & 20.0 & 21.7 & 19.6 & 20.0 & 18.9 & 20.3 & 19.1 \\
\addlinespace[2pt]
\rowcolor{tablepink}\multicolumn{10}{@{}l}{\textbf{Reasoning}}\\
\addlinespace[2pt]
\quad AGIEval Eng & 39.1 & 49.8 & 50.5 & 51.2 & 50.9 & 50.5 & 50.8 & 51.1 & 52.4 \\
\quad GPQA-D & 22.8 & 22.3 & 21.2 & 26.7 & 25.0 & 28.1 & 23.9 & 27.6 & 26.7 \\
\quad MMLU-PRO & 23.5 & 30.0 & 28.5 & 37.5 & 32.2 & 36.9 & 32.1 & 36.4 & 36.8 \\
\quad ZebraLogic & 0.2 & 13.4 & 14.6 & 18.3 & 10.1 & 17.7 & 11.0 & 19.2 & 17.1 \\
\quad BBH & 38.1 & 49.7 & 50.0 & 51.4 & 51.0 & 48.8 & 46.0 & 43.8 & 45.2 \\
\addlinespace[2pt]
\rowcolor{tablepink}\multicolumn{10}{@{}l}{\textbf{Coding}}\\
\addlinespace[2pt]
\quad HumanEval+ & 1.0 & 44.9 & 45.1 & 58.5 & 38.1 & 55.1 & 44.6 & 53.7 & 49.8 \\
\quad LiveCode v6 & 0.0 & 11.3 & 10.7 & 9.8 & 8.9 & 9.5 & 9.1 & 10.0 & 9.8 \\
\addlinespace[2pt]
\rowcolor{tablepink}\multicolumn{10}{@{}l}{\textbf{Instruction Following}}\\
\addlinespace[2pt]
\quad IFBench & 26.9 & 23.5 & 23.1 & 21.8 & 22.1 & 22.1 & 22.8 & 19.0 & 25.2 \\
\quad IFEval & 35.7 & 51.4 & 51.9 & 64.9 & 49.0 & 64.9 & 64.3 & 67.1 & 65.2 \\
\textbf{General Avg} & 20.8 & 32.9 & 32.9 & \textbf{\textit{37.8}} & 31.9 & 37.1 & 33.9 & 36.4 & 36.5 \\
\midrule
\textbf{Avg} & 16.4 & 29.6 & 28.9 & \textbf{\textit{32.8}} & 28.2 & 31.8 & 29.3 & 31.5 & 31.1 \\
\bottomrule
\end{tabularx}%
}
\caption{Performance of Softmax DPO+ on gemma-3-4b-sft with group size 8, varying $\beta$ and nll.}
\end{table*}

\begin{table*}[ht]
\centering
\scriptsize
\setlength{\tabcolsep}{1.8pt}
\renewcommand{\arraystretch}{1.20}
\resizebox{\linewidth}{!}{%
\begin{tabularx}{\linewidth}{@{}>{\raggedright\arraybackslash}Xccccccccc@{}}
\toprule
\multicolumn{10}{c}{\textbf{\textit{olmo-3-7b-it-sft}}} \\
\multirow{4}{*}{Dataset} & \multicolumn{9}{c}{DPO} \\
\cmidrule(lr){2-10}
 & \multicolumn{2}{c}{nll=0.0} & \multicolumn{7}{c}{nll=1.0} \\
\cmidrule(lr){2-3} \cmidrule(lr){4-10}
 & \multicolumn{2}{c}{$\beta$=2.0} & \multicolumn{2}{c}{$\beta$=2.0} & \multicolumn{2}{c}{$\beta$=5.0} & \multicolumn{1}{c}{$\beta$=10.0} & \multicolumn{2}{c}{$\beta$=15.0} \\
\cmidrule(lr){2-3} \cmidrule(lr){4-5} \cmidrule(lr){6-7} \cmidrule(lr){8-8} \cmidrule(lr){9-10}
 & 1100 & 1139 & 1100 & 1139 & 1000 & 1139 & 1139 & 800 & 1139 \\
\midrule
\addlinespace[2pt]
\rowcolor{tablepink}\multicolumn{10}{@{}l}{\textbf{Math}}\\
\addlinespace[2pt]
\quad AIME24 & 21.9 & 21.5 & 25.2 & 23.7 & 18.3 & 20.6 & 16.2 & 14.4 & 14.8 \\
\quad AIME25 & 20.4 & 19.6 & 25.2 & 21.3 & 20.6 & 19.0 & 13.8 & 10.6 & 10.2 \\
\quad AMC23 & 60.0 & 56.2 & 61.6 & 60.3 & 62.8 & 60.6 & 55.2 & 53.9 & 51.9 \\
\quad MATH500 & 81.2 & 80.0 & 81.0 & 85.2 & 82.4 & 81.4 & 79.4 & 76.8 & 75.8 \\
\quad Minerva & 31.6 & 30.5 & 34.6 & 34.6 & 33.5 & 35.3 & 33.8 & 34.2 & 32.0 \\
\quad Olympiad & 47.9 & 48.9 & 51.4 & 52.4 & 49.6 & 49.6 & 44.6 & 43.0 & 41.5 \\
\textbf{Math Avg} & 43.8 & 42.8 & \textbf{\textit{46.5}} & 46.3 & 44.5 & 44.4 & 40.5 & 38.8 & 37.7 \\
\addlinespace[2pt]
\rowcolor{tablepink}\multicolumn{10}{@{}l}{\textbf{Reasoning}}\\
\addlinespace[2pt]
\quad AGIEval Eng & 55.9 & 55.1 & 58.8 & 58.1 & 58.7 & 57.9 & 55.7 & 56.5 & 56.5 \\
\quad GPQA-D & 35.6 & 36.3 & 33.7 & 33.6 & 36.7 & 35.7 & 36.1 & 36.3 & 35.1 \\
\quad MMLU-PRO & 51.9 & 51.3 & 51.6 & 51.5 & 51.9 & 52.3 & 51.3 & 50.5 & 50.8 \\
\quad ZebraLogic & 17.7 & 17.3 & 25.4 & 24.4 & 25.2 & 25.4 & 22.2 & 19.6 & 20.5 \\
\quad BBH & 67.9 & 68.1 & 71.5 & 71.6 & 51.5 & 51.3 & 50.4 & 49.5 & 49.3 \\
\addlinespace[2pt]
\rowcolor{tablepink}\multicolumn{10}{@{}l}{\textbf{Coding}}\\
\addlinespace[2pt]
\quad HumanEval+ & 67.1 & 68.6 & 74.5 & 74.5 & 73.4 & 72.3 & 76.3 & 74.6 & 76.8 \\
\quad LiveCode v6 & 15.4 & 15.6 & 19.5 & 19.7 & 17.9 & 19.3 & 17.5 & 16.9 & 16.7 \\
\addlinespace[2pt]
\rowcolor{tablepink}\multicolumn{10}{@{}l}{\textbf{Instruction Following}}\\
\addlinespace[2pt]
\quad IFBench & 33.3 & 34.0 & 29.6 & 29.9 & 28.9 & 30.3 & 28.6 & 28.9 & 28.2 \\
\quad IFEval & 79.7 & 81.3 & 83.4 & 83.5 & 85.4 & 83.2 & 83.5 & 83.2 & 84.8 \\
\textbf{General Avg} & 47.2 & 47.5 & \textbf{\textit{49.8}} & 49.6 & 47.7 & 47.5 & 46.9 & 46.2 & 46.5 \\
\midrule
\textbf{Avg} & 45.8 & 45.6 & \textbf{\textit{48.5}} & 48.3 & 46.5 & 46.3 & 44.3 & 43.3 & 43.0 \\
\bottomrule
\end{tabularx}%
}
\caption{Performance of DPO on olmo-3-7b-it-sft with group size 1, varying $\beta$ and nll.}
\end{table*}

\begin{table*}[ht]
\centering
\scriptsize
\setlength{\tabcolsep}{1.8pt}
\renewcommand{\arraystretch}{1.20}
\resizebox{\linewidth}{!}{%
\begin{tabularx}{\linewidth}{@{}>{\raggedright\arraybackslash}Xcccccccc@{}}
\toprule
\multicolumn{9}{c}{\textbf{\textit{olmo-3-7b-it-sft}}} \\
\multirow{5}{*}{Dataset} & \multicolumn{2}{c}{Margin DPO+} & \multicolumn{2}{c}{MPO+} & \multicolumn{2}{c}{All-Pairs DPO+} & \multicolumn{2}{c}{Softmax DPO+} \\
\cmidrule(lr){2-3} \cmidrule(lr){4-5} \cmidrule(lr){6-7} \cmidrule(lr){8-9}
 & \multicolumn{2}{c}{nll=1.0} & \multicolumn{2}{c}{nll=1.0} & \multicolumn{2}{c}{nll=1.0} & \multicolumn{2}{c}{nll=1.0} \\
\cmidrule(lr){2-3} \cmidrule(lr){4-5} \cmidrule(lr){6-7} \cmidrule(lr){8-9}
 & \multicolumn{2}{c}{$\beta$=15.0} & \multicolumn{2}{c}{$\beta$=5.0} & \multicolumn{2}{c}{$\beta$=2.0} & \multicolumn{2}{c}{$\beta$=2.0} \\
\cmidrule(lr){2-3} \cmidrule(lr){4-5} \cmidrule(lr){6-7} \cmidrule(lr){8-9}
 & 1000 & 1139 & 1000 & 1139 & 900 & 1139 & 800 & 1139 \\
\midrule
\addlinespace[2pt]
\rowcolor{tablepink}\multicolumn{9}{@{}l}{\textbf{Math}}\\
\addlinespace[2pt]
\quad AIME24 & 18.8 & 19.4 & 28.3 & 26.9 & 28.1 & 27.5 & 28.1 & 29.2 \\
\quad AIME25 & 16.9 & 15.4 & 23.1 & 24.2 & 23.3 & 23.5 & 22.7 & 24.6 \\
\quad AMC23 & 57.5 & 58.6 & 66.9 & 66.6 & 67.5 & 62.8 & 68.4 & 67.2 \\
\quad MATH500 & 81.6 & 79.8 & 86.0 & 83.0 & 85.2 & 86.2 & 86.8 & 86.8 \\
\quad Minerva & 32.0 & 33.8 & 35.3 & 36.4 & 35.7 & 34.6 & 39.0 & 35.3 \\
\quad Olympiad & 48.9 & 46.5 & 52.4 & 54.5 & 53.2 & 52.9 & 55.6 & 55.0 \\
\textbf{Math Avg} & 42.6 & 42.3 & 48.7 & 48.6 & 48.8 & 47.9 & \textbf{\textit{50.1}} & 49.7 \\
\addlinespace[2pt]
\rowcolor{tablepink}\multicolumn{9}{@{}l}{\textbf{Reasoning}}\\
\addlinespace[2pt]
\quad AGIEval Eng & 56.4 & 57.0 & 58.9 & 59.4 & 59.5 & 60.9 & 58.5 & 60.3 \\
\quad GPQA-D & 36.4 & 36.4 & 35.7 & 36.9 & 35.2 & 36.4 & 36.0 & 35.7 \\
\quad MMLU-PRO & 51.3 & 51.8 & 53.1 & 53.0 & 53.3 & 53.5 & 53.1 & 54.0 \\
\quad ZebraLogic & 24.8 & 24.6 & 28.1 & 29.3 & 28.9 & 29.2 & 29.1 & 30.0 \\
\quad BBH & 50.1 & 50.3 & 65.7 & 66.5 & 72.2 & 73.1 & 72.4 & 73.0 \\
\addlinespace[2pt]
\rowcolor{tablepink}\multicolumn{9}{@{}l}{\textbf{Coding}}\\
\addlinespace[2pt]
\quad HumanEval+ & 74.7 & 75.9 & 71.8 & 72.0 & 79.5 & 79.5 & 75.6 & 75.7 \\
\quad LiveCode v6 & 16.9 & 17.7 & 19.9 & 19.1 & 20.3 & 20.3 & 19.0 & 19.6 \\
\addlinespace[2pt]
\rowcolor{tablepink}\multicolumn{9}{@{}l}{\textbf{Instruction Following}}\\
\addlinespace[2pt]
\quad IFBench & 29.3 & 30.3 & 29.3 & 29.6 & 33.0 & 31.3 & 31.0 & 31.0 \\
\quad IFEval & 83.9 & 83.4 & 82.8 & 83.4 & 84.7 & 84.8 & 83.7 & 85.6 \\
\textbf{General Avg} & 47.1 & 47.5 & 49.5 & 49.9 & 51.8 & \textbf{\textit{52.1}} & 50.9 & 51.7 \\
\midrule
\textbf{Avg} & 45.3 & 45.4 & 49.2 & 49.4 & 50.6 & 50.4 & 50.6 & \textbf{\textit{50.9}} \\
\bottomrule
\end{tabularx}%
}
\caption{Performance of olmo-3-7b-it-sft with group size 2, varying objectives, $\beta$ and nll.}
\end{table*}

\begin{table*}[ht]
\centering
\scriptsize
\setlength{\tabcolsep}{1.8pt}
\renewcommand{\arraystretch}{1.20}
\resizebox{\linewidth}{!}{%
\begin{tabularx}{\linewidth}{@{}>{\raggedright\arraybackslash}Xcccccc@{}}
\toprule
\multicolumn{7}{c}{\textbf{\textit{olmo-3-7b-it-sft}}} \\
\multirow{5}{*}{Dataset} & \multicolumn{2}{c}{Margin DPO+} & \multicolumn{2}{c}{MPO+} & \multicolumn{1}{c}{All-Pairs DPO+} & \multicolumn{1}{c}{Softmax DPO+} \\
\cmidrule(lr){2-3} \cmidrule(lr){4-5} \cmidrule(lr){6-6} \cmidrule(lr){7-7}
 & \multicolumn{2}{c}{nll=1.0} & \multicolumn{2}{c}{nll=1.0} & \multicolumn{1}{c}{nll=1.0} & \multicolumn{1}{c}{nll=1.0} \\
\cmidrule(lr){2-3} \cmidrule(lr){4-5} \cmidrule(lr){6-6} \cmidrule(lr){7-7}
 & \multicolumn{2}{c}{$\beta$=15.0} & \multicolumn{2}{c}{$\beta$=5.0} & \multicolumn{1}{c}{$\beta$=2.0} & \multicolumn{1}{c}{$\beta$=2.0} \\
\cmidrule(lr){2-3} \cmidrule(lr){4-5} \cmidrule(lr){6-6} \cmidrule(lr){7-7}
 & 1100 & 1139 & 1100 & 1139 & 1139 & 1139 \\
\midrule
\addlinespace[2pt]
\rowcolor{tablepink}\multicolumn{7}{@{}l}{\textbf{Math}}\\
\addlinespace[2pt]
\quad AIME24 & 27.3 & 25.2 & 30.2 & 28.3 & 30.0 & 34.6 \\
\quad AIME25 & 23.1 & 22.7 & 25.0 & 22.5 & 26.7 & 25.4 \\
\quad AMC23 & 62.7 & 61.1 & 69.2 & 69.1 & 70.3 & 72.7 \\
\quad MATH500 & 84.0 & 83.6 & 87.8 & 85.2 & 87.8 & 87.6 \\
\quad Minerva & 35.7 & 35.3 & 34.2 & 39.7 & 35.7 & 37.9 \\
\quad Olympiad & 53.6 & 55.6 & 55.3 & 55.0 & 57.2 & 56.7 \\
\textbf{Math Avg} & 47.7 & 47.2 & 50.3 & 50.0 & 51.3 & \textbf{\textit{52.5}} \\
\addlinespace[2pt]
\rowcolor{tablepink}\multicolumn{7}{@{}l}{\textbf{Reasoning}}\\
\addlinespace[2pt]
\quad AGIEval Eng & 59.1 & 60.2 & 59.6 & 59.8 & 62.0 & 61.2 \\
\quad GPQA-D & 37.2 & 37.0 & 36.5 & 35.8 & 35.4 & 36.9 \\
\quad MMLU-PRO & 54.1 & 53.3 & 54.2 & 54.3 & 54.6 & 54.5 \\
\quad ZebraLogic & 29.1 & 29.8 & 31.3 & 33.4 & 30.9 & 28.1 \\
\quad BBH & 71.9 & 72.3 & 73.2 & 72.9 & 73.5 & 74.3 \\
\addlinespace[2pt]
\rowcolor{tablepink}\multicolumn{7}{@{}l}{\textbf{Coding}}\\
\addlinespace[2pt]
\quad HumanEval+ & 73.2 & 72.6 & 72.2 & 72.5 & 76.5 & 72.0 \\
\quad LiveCode v6 & 20.1 & 19.8 & 18.9 & 19.3 & 20.3 & 18.7 \\
\addlinespace[2pt]
\rowcolor{tablepink}\multicolumn{7}{@{}l}{\textbf{Instruction Following}}\\
\addlinespace[2pt]
\quad IFBench & 31.0 & 30.3 & 29.3 & 32.0 & 31.0 & 32.3 \\
\quad IFEval & 84.8 & 85.0 & 85.2 & 83.9 & 83.9 & 84.3 \\
\textbf{General Avg} & 51.2 & 51.1 & 51.1 & 51.5 & \textbf{\textit{52.0}} & 51.4 \\
\midrule
\textbf{Avg} & 49.8 & 49.6 & 50.8 & 50.9 & 51.7 & \textbf{\textit{51.8}} \\
\bottomrule
\end{tabularx}%
}
\caption{Performance of olmo-3-7b-it-sft with group size 4, varying objectives, $\beta$ and nll.}
\end{table*}

\begin{table*}[ht]
\centering
\scriptsize
\setlength{\tabcolsep}{1.8pt}
\renewcommand{\arraystretch}{1.20}
\resizebox{\linewidth}{!}{%
\begin{tabularx}{\linewidth}{@{}>{\raggedright\arraybackslash}Xcccccccccc@{}}
\toprule
\multicolumn{11}{c}{\textbf{\textit{olmo-3-7b-it-sft}}} \\
\multirow{4}{*}{Dataset} & \multicolumn{10}{c}{Margin DPO+} \\
\cmidrule(lr){2-11}
 & \multicolumn{2}{c}{nll=0.0} & \multicolumn{8}{c}{nll=1.0} \\
\cmidrule(lr){2-3} \cmidrule(lr){4-11}
 & \multicolumn{2}{c}{$\beta$=2.0} & \multicolumn{2}{c}{$\beta$=2.0} & \multicolumn{2}{c}{$\beta$=5.0} & \multicolumn{2}{c}{$\beta$=10.0} & \multicolumn{2}{c}{$\beta$=15.0} \\
\cmidrule(lr){2-3} \cmidrule(lr){4-5} \cmidrule(lr){6-7} \cmidrule(lr){8-9} \cmidrule(lr){10-11}
 & 200 & 1139 & 500 & 1139 & 600 & 1139 & 600 & 1139 & 1100 & 1139 \\
\midrule
\addlinespace[2pt]
\rowcolor{tablepink}\multicolumn{11}{@{}l}{\textbf{Math}}\\
\addlinespace[2pt]
\quad AIME24 & 16.0 & 0.0 & 29.8 & 23.5 & 33.3 & 24.2 & 33.3 & 29.4 & 29.6 & 26.5 \\
\quad AIME25 & 13.3 & 0.0 & 25.6 & 20.8 & 25.2 & 21.3 & 25.0 & 26.7 & 28.1 & 25.2 \\
\quad AMC23 & 54.5 & 0.0 & 70.3 & 63.6 & 67.7 & 66.6 & 67.2 & 68.9 & 67.8 & 68.9 \\
\quad MATH500 & 74.2 & 0.0 & 87.4 & 85.8 & 87.2 & 85.8 & 87.4 & 86.6 & 86.6 & 87.2 \\
\quad Minerva & 28.3 & 0.0 & 36.0 & 33.5 & 35.7 & 32.4 & 39.0 & 34.9 & 38.2 & 34.6 \\
\quad Olympiad & 43.0 & 0.0 & 54.7 & 52.0 & 55.1 & 53.3 & 55.3 & 56.0 & 56.0 & 55.4 \\
\textbf{Math Avg} & 38.2 & 0.0 & 50.6 & 46.5 & 50.7 & 47.2 & \textbf{\textit{51.2}} & 50.4 & 51.1 & 49.6 \\
\addlinespace[2pt]
\rowcolor{tablepink}\multicolumn{11}{@{}l}{\textbf{Reasoning}}\\
\addlinespace[2pt]
\quad AGIEval Eng & 49.9 & 0.0 & 59.5 & 60.1 & 60.0 & 59.4 & 59.3 & 58.6 & 59.4 & 59.1 \\
\quad GPQA-D & 34.9 & 0.0 & 36.0 & 35.4 & 37.6 & 36.6 & 37.2 & 37.1 & 36.9 & 36.7 \\
\quad MMLU-PRO & 47.7 & 0.0 & 53.7 & 52.2 & 53.9 & 52.9 & 54.2 & 54.6 & 55.0 & 54.5 \\
\quad ZebraLogic & 16.9 & 0.0 & 18.8 & 24.4 & 21.9 & 24.8 & 21.9 & 26.6 & 26.4 & 27.4 \\
\quad BBH & 63.7 & 0.0 & 71.9 & 70.3 & 72.5 & 71.5 & 73.8 & 73.5 & 73.2 & 72.8 \\
\addlinespace[2pt]
\rowcolor{tablepink}\multicolumn{11}{@{}l}{\textbf{Coding}}\\
\addlinespace[2pt]
\quad HumanEval+ & 69.8 & 0.0 & 72.0 & 72.1 & 71.5 & 74.3 & 71.3 & 73.5 & 73.4 & 76.0 \\
\quad LiveCode v6 & 15.7 & 0.0 & 17.7 & 14.6 & 18.1 & 17.5 & 20.6 & 19.7 & 22.5 & 21.2 \\
\addlinespace[2pt]
\rowcolor{tablepink}\multicolumn{11}{@{}l}{\textbf{Instruction Following}}\\
\addlinespace[2pt]
\quad IFBench & 29.3 & 19.0 & 29.9 & 27.6 & 32.7 & 29.6 & 32.7 & 29.6 & 30.3 & 29.6 \\
\quad IFEval & 76.2 & 12.8 & 83.9 & 80.6 & 84.3 & 83.5 & 83.9 & 84.5 & 84.3 & 84.8 \\
\textbf{General Avg} & 44.9 & 3.5 & 49.3 & 48.6 & 50.3 & 50.0 & 50.5 & 50.8 & 51.3 & \textbf{\textit{51.3}} \\
\midrule
\textbf{Avg} & 42.2 & 2.1 & 49.8 & 47.8 & 50.4 & 48.9 & 50.8 & 50.7 & \textbf{\textit{51.2}} & 50.7 \\
\bottomrule
\end{tabularx}%
}
\caption{Performance of Margin DPO+ on olmo-3-7b-it-sft with group size 8, varying $\beta$ and nll.}
\end{table*}

\begin{table*}[ht]
\centering
\scriptsize
\setlength{\tabcolsep}{1.8pt}
\renewcommand{\arraystretch}{1.20}
\resizebox{\linewidth}{!}{%
\begin{tabularx}{\linewidth}{@{}>{\raggedright\arraybackslash}Xcccccccccc@{}}
\toprule
\multicolumn{11}{c}{\textbf{\textit{olmo-3-7b-it-sft}}} \\
\multirow{4}{*}{Dataset} & \multicolumn{10}{c}{MPO+} \\
\cmidrule(lr){2-11}
 & \multicolumn{2}{c}{nll=0.0} & \multicolumn{8}{c}{nll=1.0} \\
\cmidrule(lr){2-3} \cmidrule(lr){4-11}
 & \multicolumn{2}{c}{$\beta$=2.0} & \multicolumn{2}{c}{$\beta$=2.0} & \multicolumn{2}{c}{$\beta$=5.0} & \multicolumn{2}{c}{$\beta$=10.0} & \multicolumn{2}{c}{$\beta$=15.0} \\
\cmidrule(lr){2-3} \cmidrule(lr){4-5} \cmidrule(lr){6-7} \cmidrule(lr){8-9} \cmidrule(lr){10-11}
 & 300 & 1139 & 1000 & 1139 & 700 & 1139 & 1000 & 1139 & 1000 & 1139 \\
\midrule
\addlinespace[2pt]
\rowcolor{tablepink}\multicolumn{11}{@{}l}{\textbf{Math}}\\
\addlinespace[2pt]
\quad AIME24 & 17.9 & 0.0 & 31.9 & 33.5 & 32.9 & 32.1 & 26.5 & 30.0 & 21.5 & 19.0 \\
\quad AIME25 & 18.1 & 0.0 & 28.1 & 23.7 & 27.7 & 26.0 & 21.9 & 21.0 & 19.0 & 18.3 \\
\quad AMC23 & 57.0 & 0.0 & 70.3 & 68.6 & 71.4 & 73.1 & 66.1 & 65.0 & 60.6 & 59.1 \\
\quad MATH500 & 78.0 & 0.4 & 86.8 & 87.4 & 87.6 & 88.4 & 85.8 & 83.2 & 79.6 & 80.6 \\
\quad Minerva & 31.2 & 0.4 & 35.7 & 38.6 & 36.8 & 34.9 & 37.9 & 33.1 & 33.5 & 31.6 \\
\quad Olympiad & 46.1 & 0.0 & 56.9 & 57.5 & 57.6 & 57.3 & 52.3 & 53.8 & 51.3 & 48.9 \\
\textbf{Math Avg} & 41.4 & 0.1 & 51.6 & 51.6 & \textbf{\textit{52.3}} & 52.0 & 48.4 & 47.7 & 44.2 & 42.9 \\
\addlinespace[2pt]
\rowcolor{tablepink}\multicolumn{11}{@{}l}{\textbf{Reasoning}}\\
\addlinespace[2pt]
\quad AGIEval Eng & 54.7 & 14.5 & 62.2 & 62.7 & 60.3 & 60.9 & 59.6 & 60.1 & 57.4 & 57.5 \\
\quad GPQA-D & 35.6 & 0.0 & 37.2 & 36.8 & 36.6 & 37.3 & 37.0 & 36.5 & 34.7 & 36.6 \\
\quad MMLU-PRO & 50.5 & 2.0 & 55.3 & 55.1 & 54.7 & 54.8 & 53.1 & 53.4 & 52.7 & 53.1 \\
\quad ZebraLogic & 15.0 & 0.0 & 27.0 & 27.3 & 29.8 & 28.8 & 29.9 & 28.6 & 26.0 & 25.5 \\
\quad BBH & 62.8 & 2.1 & 73.7 & 74.1 & 73.2 & 73.6 & 41.2 & 40.6 & 48.9 & 48.2 \\
\addlinespace[2pt]
\rowcolor{tablepink}\multicolumn{11}{@{}l}{\textbf{Coding}}\\
\addlinespace[2pt]
\quad HumanEval+ & 55.5 & 0.6 & 71.5 & 72.9 & 72.6 & 73.6 & 53.7 & 52.7 & 55.1 & 56.3 \\
\quad LiveCode v6 & 11.8 & 0.0 & 16.2 & 17.4 & 19.3 & 19.5 & 18.3 & 16.2 & 17.7 & 17.0 \\
\addlinespace[2pt]
\rowcolor{tablepink}\multicolumn{11}{@{}l}{\textbf{Instruction Following}}\\
\addlinespace[2pt]
\quad IFBench & 30.6 & 16.0 & 31.6 & 31.3 & 30.3 & 31.3 & 27.9 & 29.3 & 31.6 & 32.3 \\
\quad IFEval & 76.2 & 30.9 & 80.6 & 81.7 & 83.5 & 84.1 & 86.0 & 85.0 & 83.5 & 84.5 \\
\textbf{General Avg} & 43.6 & 7.3 & 50.6 & 51.0 & 51.2 & \textbf{\textit{51.6}} & 45.2 & 44.7 & 45.3 & 45.6 \\
\midrule
\textbf{Avg} & 42.7 & 4.5 & 51.0 & 51.2 & 51.6 & \textbf{\textit{51.7}} & 46.5 & 45.9 & 44.9 & 44.6 \\
\bottomrule
\end{tabularx}%
}
\caption{Performance of MPO+ on olmo-3-7b-it-sft with group size 8, varying $\beta$ and nll.}
\end{table*}

\begin{table*}[ht]
\centering
\scriptsize
\setlength{\tabcolsep}{1.8pt}
\renewcommand{\arraystretch}{1.20}
\resizebox{\linewidth}{!}{%
\begin{tabularx}{\linewidth}{@{}>{\raggedright\arraybackslash}Xcccccccc@{}}
\toprule
\multicolumn{9}{c}{\textbf{\textit{olmo-3-7b-it-sft}}} \\
\multirow{4}{*}{Dataset} & \multicolumn{8}{c}{All-Pairs DPO+} \\
\cmidrule(lr){2-9}
 & \multicolumn{2}{c}{nll=0.0} & \multicolumn{6}{c}{nll=1.0} \\
\cmidrule(lr){2-3} \cmidrule(lr){4-9}
 & \multicolumn{2}{c}{$\beta$=2.0} & \multicolumn{1}{c}{$\beta$=2.0} & \multicolumn{2}{c}{$\beta$=5.0} & \multicolumn{1}{c}{$\beta$=10.0} & \multicolumn{2}{c}{$\beta$=15.0} \\
\cmidrule(lr){2-3} \cmidrule(lr){4-4} \cmidrule(lr){5-6} \cmidrule(lr){7-7} \cmidrule(lr){8-9}
 & 300 & 1139 & 1139 & 1100 & 1139 & 1139 & 1000 & 1139 \\
\midrule
\addlinespace[2pt]
\rowcolor{tablepink}\multicolumn{9}{@{}l}{\textbf{Math}}\\
\addlinespace[2pt]
\quad AIME24 & 19.4 & 0.0 & 32.1 & 31.2 & 28.5 & 22.5 & 18.5 & 17.9 \\
\quad AIME25 & 17.5 & 1.5 & 26.9 & 24.8 & 22.5 & 20.2 & 16.2 & 14.0 \\
\quad AMC23 & 57.5 & 4.4 & 69.5 & 71.2 & 71.6 & 63.4 & 57.7 & 55.5 \\
\quad MATH500 & 79.0 & 10.2 & 89.2 & 87.2 & 86.2 & 82.4 & 79.0 & 78.4 \\
\quad Minerva & 32.4 & 1.1 & 38.2 & 38.6 & 36.8 & 33.8 & 33.1 & 31.6 \\
\quad Olympiad & 49.9 & 1.2 & 57.3 & 56.1 & 55.1 & 51.7 & 49.0 & 46.7 \\
\textbf{Math Avg} & 42.6 & 3.1 & \textbf{\textit{52.2}} & 51.5 & 50.1 & 45.7 & 42.3 & 40.7 \\
\addlinespace[2pt]
\rowcolor{tablepink}\multicolumn{9}{@{}l}{\textbf{Reasoning}}\\
\addlinespace[2pt]
\quad AGIEval Eng & 55.9 & 20.5 & 61.8 & 61.3 & 60.9 & 58.2 & 56.5 & 57.1 \\
\quad GPQA-D & 36.2 & 0.6 & 36.8 & 36.1 & 37.2 & 37.3 & 36.8 & 36.4 \\
\quad MMLU-PRO & 52.5 & 13.2 & 55.2 & 55.2 & 54.8 & 53.2 & 51.5 & 52.0 \\
\quad ZebraLogic & 16.8 & 6.4 & 30.9 & 34.1 & 36.0 & 27.8 & 23.4 & 24.3 \\
\quad BBH & 68.2 & 18.3 & 73.8 & 53.5 & 53.8 & 50.6 & 50.9 & 50.2 \\
\addlinespace[2pt]
\rowcolor{tablepink}\multicolumn{9}{@{}l}{\textbf{Coding}}\\
\addlinespace[2pt]
\quad HumanEval+ & 71.6 & 7.2 & 73.2 & 72.5 & 73.7 & 76.2 & 77.1 & 74.9 \\
\quad LiveCode v6 & 17.1 & 2.1 & 18.0 & 23.0 & 21.9 & 21.1 & 18.1 & 18.4 \\
\addlinespace[2pt]
\rowcolor{tablepink}\multicolumn{9}{@{}l}{\textbf{Instruction Following}}\\
\addlinespace[2pt]
\quad IFBench & 30.3 & 24.1 & 34.4 & 30.3 & 29.6 & 29.6 & 29.9 & 29.6 \\
\quad IFEval & 81.0 & 59.1 & 82.1 & 84.8 & 85.4 & 86.0 & 85.4 & 84.7 \\
\textbf{General Avg} & 47.7 & 16.9 & \textbf{\textit{51.8}} & 50.1 & 50.4 & 48.9 & 47.7 & 47.5 \\
\midrule
\textbf{Avg} & 45.7 & 11.3 & \textbf{\textit{52.0}} & 50.7 & 50.3 & 47.6 & 45.6 & 44.8 \\
\bottomrule
\end{tabularx}%
}
\caption{Performance of All-Pairs DPO+ on olmo-3-7b-it-sft with group size 8, varying $\beta$ and nll.}
\end{table*}

\begin{table*}[ht]
\centering
\scriptsize
\setlength{\tabcolsep}{1.8pt}
\renewcommand{\arraystretch}{1.20}
\resizebox{\linewidth}{!}{%
\begin{tabularx}{\linewidth}{@{}>{\raggedright\arraybackslash}Xcccccccccc@{}}
\toprule
\multicolumn{11}{c}{\textbf{\textit{olmo-3-7b-it-sft}}} \\
\multirow{4}{*}{Dataset} & \multicolumn{10}{c}{Softmax DPO+} \\
\cmidrule(lr){2-11}
 & \multicolumn{2}{c}{nll=0.0} & \multicolumn{8}{c}{nll=1.0} \\
\cmidrule(lr){2-3} \cmidrule(lr){4-11}
 & \multicolumn{2}{c}{$\beta$=2.0} & \multicolumn{2}{c}{$\beta$=2.0} & \multicolumn{2}{c}{$\beta$=5.0} & \multicolumn{2}{c}{$\beta$=10.0} & \multicolumn{2}{c}{$\beta$=15.0} \\
\cmidrule(lr){2-3} \cmidrule(lr){4-5} \cmidrule(lr){6-7} \cmidrule(lr){8-9} \cmidrule(lr){10-11}
 & 300 & 1139 & 900 & 1139 & 1000 & 1139 & 1000 & 1139 & 1100 & 1139 \\
\midrule
\addlinespace[2pt]
\rowcolor{tablepink}\multicolumn{11}{@{}l}{\textbf{Math}}\\
\addlinespace[2pt]
\quad AIME24 & 19.6 & 0.0 & 31.9 & 35.2 & 28.7 & 27.9 & 20.4 & 18.3 & 15.8 & 14.8 \\
\quad AIME25 & 17.9 & 0.8 & 26.5 & 25.2 & 21.3 & 21.7 & 15.8 & 15.6 & 10.8 & 11.2 \\
\quad AMC23 & 58.1 & 1.6 & 72.8 & 71.6 & 71.6 & 69.2 & 57.7 & 55.0 & 54.4 & 52.7 \\
\quad MATH500 & 79.6 & 7.8 & 87.8 & 86.0 & 85.8 & 85.8 & 79.4 & 79.6 & 77.2 & 76.6 \\
\quad Minerva & 32.7 & 1.1 & 39.3 & 34.2 & 35.7 & 33.5 & 33.1 & 30.9 & 33.1 & 32.7 \\
\quad Olympiad & 49.6 & 1.0 & 57.3 & 56.6 & 53.5 & 53.9 & 47.0 & 47.9 & 43.3 & 42.8 \\
\textbf{Math Avg} & 42.9 & 2.1 & \textbf{\textit{52.6}} & 51.5 & 49.4 & 48.7 & 42.2 & 41.2 & 39.1 & 38.5 \\
\addlinespace[2pt]
\rowcolor{tablepink}\multicolumn{11}{@{}l}{\textbf{Reasoning}}\\
\addlinespace[2pt]
\quad AGIEval Eng & 55.8 & 12.4 & 61.4 & 61.0 & 60.9 & 60.9 & 57.6 & 58.0 & 57.1 & 57.6 \\
\quad GPQA-D & 36.7 & 0.1 & 35.4 & 34.9 & 38.2 & 36.4 & 36.6 & 37.3 & 35.5 & 36.0 \\
\quad MMLU-PRO & 52.3 & 9.6 & 53.6 & 53.0 & 54.6 & 54.4 & 53.1 & 53.3 & 51.7 & 51.9 \\
\quad ZebraLogic & 19.8 & 4.7 & 25.5 & 26.1 & 31.6 & 31.3 & 26.6 & 26.1 & 22.7 & 21.4 \\
\quad BBH & 68.0 & 14.9 & 72.9 & 72.9 & 41.2 & 42.1 & 49.6 & 49.3 & 49.4 & 48.9 \\
\addlinespace[2pt]
\rowcolor{tablepink}\multicolumn{11}{@{}l}{\textbf{Coding}}\\
\addlinespace[2pt]
\quad HumanEval+ & 72.3 & 3.9 & 65.9 & 62.9 & 64.9 & 65.2 & 71.8 & 71.3 & 74.8 & 74.6 \\
\quad LiveCode v6 & 16.4 & 2.1 & 16.1 & 15.5 & 19.0 & 19.3 & 16.9 & 17.4 & 16.5 & 17.4 \\
\addlinespace[2pt]
\rowcolor{tablepink}\multicolumn{11}{@{}l}{\textbf{Instruction Following}}\\
\addlinespace[2pt]
\quad IFBench & 30.6 & 20.4 & 32.7 & 31.0 & 30.6 & 30.6 & 28.2 & 32.0 & 29.6 & 28.9 \\
\quad IFEval & 81.5 & 52.3 & 83.9 & 83.2 & 85.0 & 83.7 & 85.6 & 85.4 & 85.2 & 84.8 \\
\textbf{General Avg} & 48.2 & 13.4 & \textbf{\textit{49.7}} & 48.9 & 47.3 & 47.1 & 47.3 & 47.8 & 46.9 & 46.8 \\
\midrule
\textbf{Avg} & 46.1 & 8.8 & \textbf{\textit{50.9}} & 50.0 & 48.2 & 47.7 & 45.3 & 45.2 & 43.8 & 43.5 \\
\bottomrule
\end{tabularx}%
}
\caption{Performance of Softmax DPO+ on olmo-3-7b-it-sft with group size 8, varying $\beta$ and nll.}
\end{table*}

\begin{table*}[ht]
\centering
\scriptsize
\setlength{\tabcolsep}{1.8pt}
\renewcommand{\arraystretch}{1.20}
\resizebox{\linewidth}{!}{%
\begin{tabularx}{\linewidth}{@{}>{\raggedright\arraybackslash}Xcccccccccc@{}}
\toprule
\multicolumn{11}{c}{\textbf{\textit{olmo-3.1-32b-it-sft}}} \\
\multirow{5}{*}{Dataset} & \multicolumn{2}{c}{DPO} & \multicolumn{2}{c}{Margin DPO+} & \multicolumn{2}{c}{MPO+} & \multicolumn{2}{c}{All-Pairs DPO+} & \multicolumn{2}{c}{Softmax DPO+} \\
\cmidrule(lr){2-3} \cmidrule(lr){4-5} \cmidrule(lr){6-7} \cmidrule(lr){8-9} \cmidrule(lr){10-11}
 & \multicolumn{2}{c}{nll=1.0} & \multicolumn{2}{c}{nll=1.0} & \multicolumn{2}{c}{nll=1.0} & \multicolumn{2}{c}{nll=1.0} & \multicolumn{2}{c}{nll=1.0} \\
\cmidrule(lr){2-3} \cmidrule(lr){4-5} \cmidrule(lr){6-7} \cmidrule(lr){8-9} \cmidrule(lr){10-11}
 & \multicolumn{2}{c}{$\beta$=2.0} & \multicolumn{2}{c}{$\beta$=2.0} & \multicolumn{2}{c}{$\beta$=2.0} & \multicolumn{2}{c}{$\beta$=2.0} & \multicolumn{2}{c}{$\beta$=2.0} \\
\cmidrule(lr){2-3} \cmidrule(lr){4-5} \cmidrule(lr){6-7} \cmidrule(lr){8-9} \cmidrule(lr){10-11}
 & 1000 & 1139 & 600 & 1139 & 1000 & 1139 & 800 & 1139 & 800 & 1139 \\
\midrule
\addlinespace[2pt]
\rowcolor{tablepink}\multicolumn{11}{@{}l}{\textbf{Math}}\\
\addlinespace[2pt]
\quad AIME24 & 32.1 & 30.0 & 40.8 & 40.0 & 46.0 & 43.8 & 46.7 & 42.9 & 44.2 & 43.8 \\
\quad AIME25 & 25.6 & 21.7 & 33.3 & 30.6 & 35.8 & 34.4 & 31.7 & 35.2 & 32.3 & 32.7 \\
\quad AMC23 & 67.7 & 68.6 & 76.6 & 77.7 & 78.3 & 78.0 & 77.2 & 77.2 & 77.5 & 77.3 \\
\quad MATH500 & 86.6 & 86.2 & 90.6 & 89.6 & 90.0 & 91.8 & 90.8 & 90.2 & 89.8 & 89.8 \\
\quad Minerva & 45.2 & 43.4 & 43.8 & 38.2 & 43.8 & 42.3 & 44.5 & 43.0 & 43.4 & 43.4 \\
\quad Olympiad & 59.9 & 58.7 & 61.3 & 62.7 & 64.7 & 65.3 & 67.0 & 62.5 & 64.1 & 63.1 \\
\textbf{Math Avg} & 52.8 & 51.4 & 57.7 & 56.5 & \textbf{\textit{59.8}} & 59.3 & 59.6 & 58.5 & 58.5 & 58.3 \\
\addlinespace[2pt]
\rowcolor{tablepink}\multicolumn{11}{@{}l}{\textbf{Reasoning}}\\
\addlinespace[2pt]
\quad AGIEval Eng & 73.6 & 73.8 & 72.6 & 72.0 & 74.3 & 74.5 & 74.4 & 74.4 & 74.4 & 74.4 \\
\quad GPQA-D & 47.8 & 47.5 & 46.6 & 46.0 & 48.5 & 47.9 & 48.9 & 48.2 & 49.1 & 48.9 \\
\quad MMLU-PRO & 67.9 & 68.2 & 67.3 & 66.7 & 69.3 & 69.4 & 69.4 & 69.3 & 69.1 & 68.7 \\
\quad ZebraLogic & 19.1 & 17.7 & 12.1 & 15.8 & 18.2 & 17.0 & 21.5 & 17.9 & 18.6 & 16.7 \\
\quad BBH & 81.6 & 81.6 & 81.6 & 81.0 & 82.3 & 82.3 & 82.1 & 82.3 & 81.6 & 81.3 \\
\addlinespace[2pt]
\rowcolor{tablepink}\multicolumn{11}{@{}l}{\textbf{Coding}}\\
\addlinespace[2pt]
\quad HumanEval+ & 88.5 & 88.6 & 84.8 & 84.2 & 85.3 & 86.6 & 87.6 & 86.7 & 84.5 & 84.9 \\
\quad LiveCode v6 & 29.0 & 30.1 & 28.6 & 25.7 & 26.6 & 25.9 & 29.7 & 29.6 & 27.5 & 27.0 \\
\addlinespace[2pt]
\rowcolor{tablepink}\multicolumn{11}{@{}l}{\textbf{Instruction Following}}\\
\addlinespace[2pt]
\quad IFBench & 36.1 & 36.4 & 35.4 & 35.7 & 33.3 & 34.4 & 35.0 & 35.4 & 34.4 & 36.4 \\
\quad IFEval & 87.1 & 87.8 & 86.7 & 85.8 & 85.8 & 86.0 & 87.2 & 88.5 & 85.6 & 86.7 \\
\textbf{General Avg} & 59.0 & 59.1 & 57.3 & 57.0 & 58.2 & 58.2 & \textbf{\textit{59.5}} & 59.1 & 58.3 & 58.3 \\
\midrule
\textbf{Avg} & 56.5 & 56.0 & 57.5 & 56.8 & 58.8 & 58.6 & \textbf{\textit{59.6}} & 58.9 & 58.4 & 58.3 \\
\bottomrule
\end{tabularx}%
}
\caption{Performance of olmo-3.1-32b-it-sft, varying objectives, $\beta$ and nll.}
\label{fig:full_results_end}
\end{table*}

\end{document}